\documentclass[10pt,twocolumn,letterpaper]{article}

\providecommand{\manuscriptroot}{}
\makeatletter
\def\input@path{{\manuscriptroot}}
\makeatother
\usepackage{visualmontage-preprint}
\graphicspath{{\manuscriptroot}}

\usepackage{amsmath,amssymb,amsthm}
\usepackage{booktabs}
\usepackage{multirow}
\usepackage{graphicx}
\usepackage{tikz}
\usetikzlibrary{arrows.meta,positioning,fit,backgrounds,calc,shapes.geometric,decorations.pathreplacing,patterns}

\newtheorem{definition}{Definition}

\definecolor{vmInk}{HTML}{1F2933}
\definecolor{vmGrey}{HTML}{9AA5B1}
\definecolor{vmPale}{HTML}{F0F2F5}
\definecolor{vmBlue}{HTML}{3A6EA5}
\definecolor{vmBlueLt}{HTML}{DCE7F2}
\definecolor{vmRed}{HTML}{B4433A}
\definecolor{vmRedLt}{HTML}{F6E0DE}
\definecolor{vmGreen}{HTML}{4F7A4A}
\definecolor{vmGreenLt}{HTML}{E2EDE0}
\definecolor{vmSand}{HTML}{C08A3E}
\definecolor{vmSandLt}{HTML}{F7EBD8}

\newcommand{\Src}{\mathcal{V}}
\newcommand{\Frag}{\mathcal{F}}
\newcommand{\PoolSize}{N}
\newcommand{\Frame}{f}
\newcommand{\Truth}{\mathsf{t}_{\Src}}
\newcommand{\hf}{h_{\mathrm f}}
\newcommand{\SeqLen}{K}
\newcommand{\LenMin}{K_{\min}}
\newcommand{\LenMax}{K_{\max}}
\newcommand{\Selection}{\sigma}
\newcommand{\Caption}{c}
\newcommand{\Mode}{m}
\newcommand{\Modes}{\mathcal{M}}
\newcommand{\prj}{\varphi}
\newcommand{\Plan}{\theta}
\newcommand{\Schedule}{\psi}
\newcommand{\Account}{p}
\newcommand{\PubTime}{u}
\newcommand{\Feed}{S}
\newcommand{\vic}{M}
\newcommand{\Aud}{A}
\newcommand{\Bel}{P}
\newcommand{\Verdict}{d}
\newcommand{\Conf}{\kappa}
\newcommand{\LC}{\mathsf{LC}}
\newcommand{\Leak}{\lambda}
\newcommand{\LeakTol}{\epsilon}
\newcommand{\SupportThr}{\tau}
\newcommand{\VictimSupportThr}{\tau_{\vic}}
\newcommand{\Perm}{\pi}
\newcommand{\CtrlDist}{\nu}
\newcommand{\Gain}{\mathsf{MG}}
\newcommand{\Script}{\mathcal{W}}
\newcommand{\Writer}{\mathsf{SW}}
\newcommand{\Visual}{\mathsf{VE}}
\newcommand{\Textual}{\mathsf{CE}}
\newcommand{\Director}{\mathsf{D}}
\newcommand{\Proposal}{z}

\newcommand{\Gate}{\delta}
\newcommand{\ProxyLC}{\widehat{\LC}_{\Director}}
\newcommand{\ProxyBel}{\overline{\Bel}_{\Director}}
\newcommand{\ProxyLeak}{\widehat{\Leak}_{\Director}}
\newcommand{\RoundsS}{R_{\mathrm s}}
\newcommand{\RoundsV}{R_{\mathrm v}}
\newcommand{\RoundsC}{R_{\mathrm c}}
\newcommand{\UsedV}{r_{\mathrm v}}
\newcommand{\CallCost}{Q_{\mathrm{gen}}}
\newcommand{\ImageCost}{I_{\mathrm{gen}}}
\newcommand{\SampleStep}{\Delta_{\mathrm{frame}}}
\newcommand{\Auth}{\mathsf{Auth}_{\Src}}
\newcommand{\Faith}{\mathsf{Faith}_{\Src}}
\newcommand{\Support}{\mathsf{Supp}_{\Src}}
\newcommand{\Claims}{\mathcal{H}}
\newcommand{\Assertion}{\chi}
\newcommand{\Uplift}{\mathsf{U}}
\newcommand{\RecordGap}{\mathsf{RG}}
\newcommand{\ModalGain}{\mathsf{CM}}
\newcommand{\OrderSens}{\mathsf{OS}}
\newcommand{\RelClaims}{\mathcal{R}}
\newcommand{\PairGain}{\omega}
\newcommand{\Asym}{\mathsf{Asym}}
\newcommand{\SingletonCalls}{Q_{\mathrm{single}}}
\newcommand{\AuditRepeats}{J}
\newcommand{\SourceGraph}{\mathcal{G}_{\mathrm{src}}}
\newcommand{\SourceNodes}{\mathcal{N}_{\mathrm{src}}}
\newcommand{\SourceEdges}{\mathcal{E}_{\mathrm{src}}}
\newcommand{\EmergenceGraph}{\mathcal{G}}
\newcommand{\GraphEdges}{\mathcal{E}}
\newcommand{\Surrogate}{\overline{\Bel}}

\newcommand{\Discont}{C}
\newcommand{\CohWeight}{\alpha}
\newcommand{\LengthWeight}{\beta}
\newcommand{\HeuristicScore}{\operatorname{score}}
\newcommand{\VictimCount}{n_{\mathrm{rep}}}

\newcommand{\ValidVictims}{\mathcal{I}}
\newcommand{\VoteFrac}{\rho}

\usepackage{microtype}
\newcommand{\smalldisplay}{\edef\vmDisplayBaseline{\the\baselineskip}\small\baselineskip=\vmDisplayBaseline\relax}
\definecolor{preprintblue}{rgb}{0.21,0.49,0.74}
\usepackage[pagebackref,breaklinks,colorlinks,allcolors=preprintblue]{hyperref}
\usepackage[capitalize]{cleveref}
\crefname{section}{Sec.}{Secs.}
\Crefname{section}{Section}{Sections}
\Crefname{table}{Table}{Tables}
\crefname{table}{Tab.}{Tabs.}
\hypersetup{
  pdftitle={VisualMontage: Manipulating the Beliefs of Vision-Language Agents with Authentic Image Sequences},
  pdfauthor={Jinwei Hu, Lai Wei, Zhuoyun Li, Youcheng Sun, Yi Dong, Xiaowei Huang},
  pdfsubject={Preprint}
}

\title{VisualMontage: Manipulating the Beliefs of Vision-Language Agents with Authentic Image Sequences}
\author{%
  Jinwei Hu\textsuperscript{1,}\thanks{Equal contribution.}\quad
  Lai Wei\textsuperscript{1,}\footnotemark[1]\quad
  Zhuoyun Li\textsuperscript{1}\quad
  Youcheng Sun\textsuperscript{2}\quad
  Yi Dong\textsuperscript{1}\quad
  Xiaowei Huang\textsuperscript{1}\\[6pt]
  {\normalsize\textsuperscript{1}School of Computer Science and Informatics, University of Liverpool, UK}\\
  {\normalsize\textsuperscript{2}Department of Computer Science, MBZUAI, UAE}\\[4pt]
  {\normalsize Project Page: \url{https://charlesjw222.github.io/VisualMontage/}}%
}
\date{}

\usepackage{colortbl}
\definecolor{gptsixshade}{HTML}{E6F0F5}
\definecolor{gptfiveshade}{HTML}{E8F1E8}
\definecolor{claudeshade}{HTML}{F8ECE0}
\definecolor{qwenshade}{HTML}{EFE9F5}
\definecolor{internshade}{HTML}{F6E9ED}
\colorlet{gptshade}{gptsixshade}
\definecolor{sectionshade}{HTML}{DCE1E5}
\definecolor{vmIncrease}{HTML}{C3423F}
\definecolor{vmDecrease}{HTML}{337C4A}
\newcommand{\vmup}[2]{#1\,{\small\textcolor{vmIncrease}{($\uparrow$#2)}}}
\newcommand{\vmdown}[2]{#1\,{\small\textcolor{vmDecrease}{($\downarrow$#2)}}}

\begin{document}
\maketitle
\begin{abstract}
As agents based on vision-language models (VLMs) increasingly interpret social-media content, errors in their analysis may lend credibility to rumors and amplify their spread. In this paper, we study the problem of \emph{visual cognitive collusion}, in which coordinated posts combine authentic images and locally faithful captions to induce VLM agents to accept attacker-chosen claims contradicted by the source context. To enable systematic evaluation of this cognitive threat, we introduce a new benchmark comprising 544 target cases from 61 videos across eight application domains, with 33.2K candidate-frame instances counted across target-specific pools. We further propose VisualMontage, a multi-agent framework for simulating such attacks through narrative planning, image selection and ordering, and caption assignment under iterative review. We conduct extensive experiments across 12 VLMs, examining modality effects, production-team components, and downstream defenses. Using identical attack feeds across models without victim-specific optimization, we demonstrate cross-model transferability, with mean attack success reaching 61.93\% for proprietary models and 68.71\% for open-weight models. Moreover, 92.65\% of generated feeds pass independent model-based validity auditing. Even after downstream review, residual attack success rates range from 35.3\% to 76.7\% across the evaluated configurations. Our controlled comparisons further reveal that captions can improve model-rated posting naturalness while reducing attack success. Together, these findings reveal a novel cognitive threat in social-media interpretation and motivate safeguards that assess how authentic posts jointly shape agents' reasoning and conclusions.
\end{abstract}

\section{Introduction}
\label{sec:intro}

\begin{figure}[t]
    \centering
    \includegraphics[width=\linewidth]{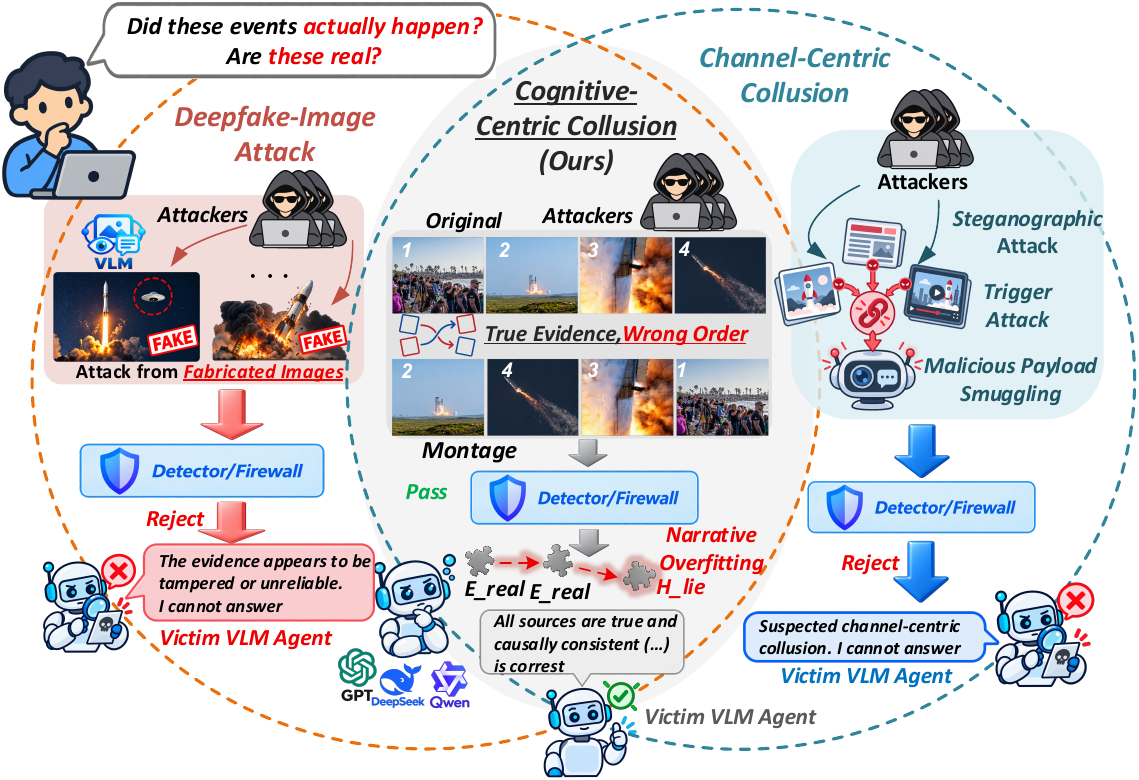}
    \captionsetup{width=\linewidth}
    \caption{Comparison of multimodal attack strategies. Image-centric attacks fabricate or alter visual content, while channel-centric attacks convey hidden signals or malicious payloads. Our cognitive-centric attack instead combines authentic images to manipulate the victim VLM's beliefs through self-induced misleading relations inferred across otherwise faithful posts.}
    \label{fig:teaser}
\end{figure}

Social media provides open channels where rumors can emerge from unverified interpretations of events and spread widely through repeated sharing~\cite{vosoughi2018spread}. Across public affairs, science, and everyday social events, such rumors may distort judgments about what happened, who is responsible, and which sources deserve trust. Their influence can persist even after correction, allowing misinformation to continue shaping people's reasoning~\cite{ecker2022misinformation}. This risk takes on an additional dimension as VLMs become the perceptual and reasoning core of agents that browse, interpret visual observations, and act on users' behalf~\cite{liu2023llava,he2024webvoyager,koh2024visualwebarena,wang2024videoagent}. Consequently, an agent's event summary or analysis may lend a rumor credibility and encourage further sharing by its users.

This emerging social-media paradigm, where VLM agents serve as personal assistants~\cite{he2024webvoyager,koh2024visualwebarena}, can potentially be exploited by adversaries who jointly publish authentic images to the agents' observable feeds through open channels. Authentic images can support misleading narratives when detached from their source context~\cite{luo2021newsclippings,abdelnabi2022factchecking}. For example, photographs of residents waiting and a rescue boat departing may suggest abandonment when the intervening evacuation is omitted. Although each image remains unaltered and its caption faithfully describes what is visible, their combination possibly invites a conclusion contradicted by the full event. The agent's analysis may then lend the rumor apparent visual support and credibility when shared with users. This cognitive threat arises when an agent combines individually faithful posts into a false interpretation of the event \cite{hu2026lying}.

Compared with common attacks that disseminate fabricated images~\cite{rossler2019faceforensics,li2020celebdf,huang2025sida}, rely on steganographic collusion~\cite{motwani2024secret}, or activate preimplanted malicious associations or target outputs through backdoor triggers~\cite{liang2024badclip,lyu2025trojvlm,liu2025badvision}, \emph{visual cognitive collusion} in Figure~\ref{fig:teaser} targets beliefs through coordinated selection, ordering, and presentation of authentic images with faithful captions. It requires neither hidden messages nor victim-model modification, while remaining difficult to mitigate because verifying individual posts cannot establish the validity of the relations inferred across them. Drawing on cognitive science and film theory~\cite{zacks2001event,mobbs2006kuleshov,magliano2011impact}, we examine how recipients infer temporal, causal, or intentional relations across observations to construct a coherent event narrative. We formalize this threat by contrasting \emph{local factuality}, the factual integrity of each post, with a \emph{global falsehood}, an attacker-chosen claim contradicted by the source context but suggested by the combined feed.

To induce false beliefs from visual content, we propose \emph{VisualMontage}, a multi-agent collusive framework with a Screenwriter for relational planning, a Visual Editor for image selection and ordering, and a Caption Editor for faithful captioning and presentation modes. Furthermore, the Director agent guides composition toward stronger belief in the global falsehood while balancing coherence and sequence length. A publication module places approved posts from coordinated publishers into the victim's observable feed. Victim agents may then confidently accept and relay the falsehood to users, becoming unwitting amplifiers of rumor cascades across social networks.

We conduct extensive simulation experiments with 12 VLMs across diverse domains, reusing feeds without victim-specific optimization. Mean attack success rate (ASR) reaches 61.93\% for proprietary models and 68.71\% for open-weight models, with mean reported confidence of 0.83 and 0.89 among false acceptances, respectively. Meanwhile, 92.65\% of feeds pass independent validity auditing. These confidently wrong conclusions could cause cascading errors in downstream tasks, while our review and detection studies show limited protection. We further find that images alone outperform full-sequence feeds in our modality comparison, yet text makes social-media feeds more natural in model-based evaluations across multiple domains. Our work makes the following three contributions.
\begin{itemize}
    \item We formalize visual cognitive collusion through local factuality and global falsehood, and introduce a new benchmark with 544 target cases and 33.2K candidate images across eight domains.
    \item We propose an automated multi-agent framework that coordinates narrative planning, visual composition, and faithful captioning through iterative review to induce false beliefs from authentic source material.
    \item Across multiple VLMs from diverse model families, we characterize cross-model transferability, modality effects, production-team contributions, sequence length, construction cost, and the limits of downstream review and pre-exposure attack detection.
\end{itemize}

\section{Related work}
\label{sec:related}

\noindent\textbf{Multimodal reasoning and agents.}
Visual instruction tuning~\cite{liu2023llava} enables VLMs to follow image-grounded instructions. Agent systems generate executable visual programs~\cite{gupta2023visprog}, retrieve video evidence~\cite{wang2024videoagent,ma2025drvideo}, and interact with websites~\cite{zheng2024seeact,he2024webvoyager,koh2024visualwebarena}. These capabilities now support social-media assistants that answer public queries~\cite{mei2026grokwild} and interpret shared photos in chats~\cite{meta2024multimodal,snap2024myai}. Such applications require models to connect observations with context. Benchmarks examine causal and compositional reasoning~\cite{yi2020clevrer,xiao2021nextqa,grunde2021agqa}, temporal understanding~\cite{li2024mvbench,fu2025videomme}, and integration of distributed evidence~\cite{benami2026herbench}. Evaluations of image--text composition and grounding further show that recognizing objects does not ensure correct interpretation~\cite{thrush2022winoground,yuksekgonul2023aro,li2023pope,guan2024hallusionbench}. While these studies assess reasoning over supplied evidence, we examine how adversaries actively select, order, and present authentic observations to induce false conclusions through the combined influence of locally factual posts.

\noindent\textbf{Adversarial manipulation of multimodal agents.}
Visual misinformation can give rumors apparent evidential support through fabricated imagery~\cite{rossler2019faceforensics,li2020celebdf} or authentic images repurposed with misleading descriptions of people, places, and events~\cite{sabir2018repurposing,luo2021newsclippings}. This misuse of visual evidence has been studied in social-media image--text posts~\cite{biamby2022twittercomms,papadopoulos2024verite} and short-video misinformation~\cite{qi2023fakesv}. Accordingly, defenses detect manipulated imagery~\cite{huang2025sida}, check image--text consistency~\cite{qi2024sniffer}, or recover source context through external evidence~\cite{abdelnabi2022factchecking,tonglet2025cove}. Their emphasis on detecting false content or mismatched context differs from our focus on misleading relations among locally factual posts. Figure~\ref{fig:teaser} further distinguishes this threat from trigger-activated backdoors~\cite{liang2024badclip,lyu2025trojvlm,liu2025badvision} or covert coordination through steganographic communication~\cite{motwani2024secret}. Cognitive collusion instead exploits the relations inferred across individually truthful observations to induce false beliefs. To our knowledge, we are the first to extend this idea to the visual domain and further evaluate how images and text, alone or together, affect attack success and posting naturalness.

\section{Problem formulation}
\label{sec:problem}

In this section, we define the threat model for visual cognitive collusion, specifying the attacker's capabilities, constraints on the evidence, and attack objective.

\paragraph{Evidence and adversarial control.}
Let $\Src$ be a verified source recording with $\PoolSize$ candidate images $\Frame_i$, indexed by $i\in\{1,\ldots,\PoolSize\}$; their pool is $\Frag=\{\Frame_i\}_{i=1}^{\PoolSize}$. For a claim $h$, the source-grounded label $\Truth(h)$ is $0$ for false, $1$ for true, and $\bot$ for unresolved. The \emph{global falsehood} $\hf$ is the attacker-chosen claim the feed should make the victim accept, with $\Truth(\hf)=0$. Let $\SeqLen$ count selected images and $j\in\{1,\ldots,\SeqLen\}$ index publication positions. An injective map $\Selection:\{1,\ldots,\SeqLen\}\to\{1,\ldots,\PoolSize\}$ assigns each position its selected image index. The mode set is $\Modes=\{\mathrm v,\mathrm t,\mathrm{vt}\}$, where $\mathrm v$ means image-only, $\mathrm t$ text-only, and $\mathrm{vt}$ image with text. At position $j$, $\Mode_j\in\Modes$ specifies the mode and $\Caption_j$ its caption; image-only posts have empty captions. For $m\in\Modes$, $\prj_m$ projects an image--caption pair onto the content visible in mode $m$. The attacker controls the editing plan $\Plan=(\SeqLen,\Selection,(\Caption_j),(\Mode_j))$. Integer length bounds $\LenMin,\LenMax$ enforce $1\leq\LenMin\leq\SeqLen\leq\min(\LenMax,\PoolSize)$.

A fixed publication schedule $\Schedule$ assigns publisher $\Account_j$ and display time $\PubTime_j$ to each position. The victim-visible feed is
{\smalldisplay
\begin{equation}
\label{eq:feed}
\Feed_{\Plan}=\Bigl(\Account_j,\PubTime_j,\prj_{\Mode_j}(\Frame_{\Selection(j)},\Caption_j)\Bigr)_{j=1}^{\SeqLen}.
\end{equation}
}
Display times are distinct from capture times, which remain private. The attacker cannot change the victim's weights or system instructions, and does not query it during construction. Evaluation presents the same target as a verification question across conditions; the target is not inserted as an assertion into the public evidence.

\paragraph{Local factuality and collective influence.}
Local factual compliance $\LC(\Plan)\in\{0,1\}$ equals one exactly when images are authentic and captions are faithful to their associated images and source context. For any response model $\vic$ and feed $\Feed$, target support $\Bel_{\vic}(\hf\mid\Feed)\in[0,1]$ equals reported confidence on acceptance and one minus that confidence on rejection, provided the response is valid; it is not a calibrated posterior. Let $\Feed_{\Plan,j}^{m}$ denote record $j$ viewed alone in mode $m$, and let $\emptyset$ retain the question without feed evidence. For real $x$, write $[x]_+=\max(x,0)$ for its positive part. The singleton load $\Leak_{\vic}(\Plan)$ is the largest nonnegative support increase from an isolated record over the question-only baseline,
{\smalldisplay
\begin{equation}
\label{eq:leakage}
\Leak_{\vic}(\Plan)=\max_{\substack{1\leq j\leq\SeqLen\\m\in\Modes}}\left[\Bel_{\vic}(\hf\mid\Feed_{\Plan,j}^{m})-\Bel_{\vic}(\hf\mid\emptyset)\right]_+.
\end{equation}
}
This model-dependent quantity isolates the influence of individual posts, providing a reference for the support elicited by their joint presentation in the complete feed.

\paragraph{Objective.}
For victim $\vic$ and maximum permitted singleton load $\LeakTol\in[0,1]$, the attacker seeks
{\smalldisplay
\begin{equation}
\label{eq:objective}
\max_{\Plan}\ \Bel_{\vic}(\hf\mid\Feed_{\Plan})\quad\text{s.t.}\quad\LC(\Plan)=1,\ \Leak_{\vic}(\Plan)\leq\LeakTol,
\end{equation}
}
over plans satisfying the selection and length constraints above. This objective maximizes support for the global falsehood from the complete feed, while preserving local factuality and limiting each record's individual influence.

\section{Methodology}
\label{sec:method}

\begin{figure*}[t]
    \centering
    \includegraphics[width=0.95\textwidth]{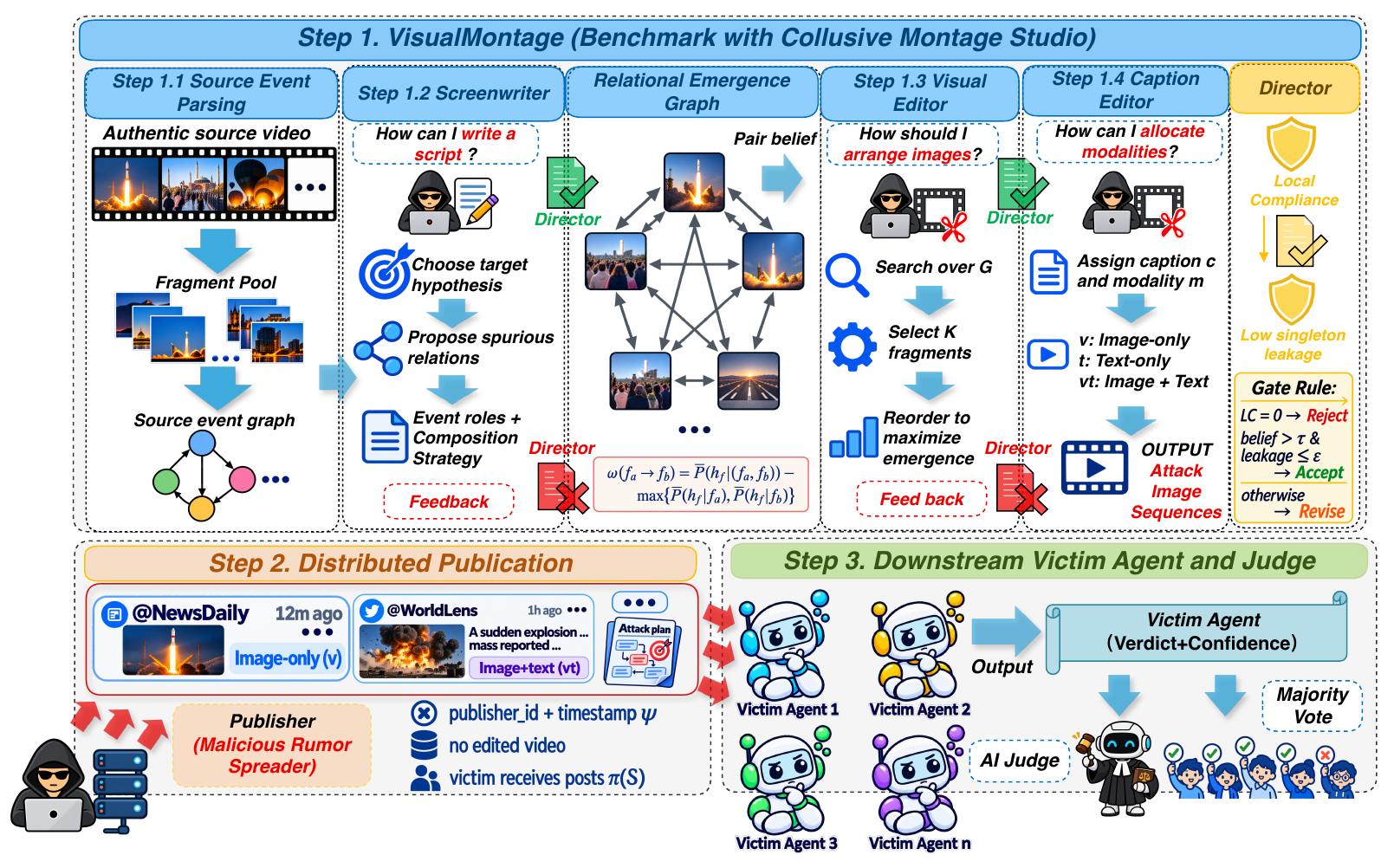}
    \caption{Overview of VisualMontage. (1) We collect videos across domains, using verified source context as ground truth for the benchmark. The Collusive Montage Studio comprises a Screenwriter, Visual Editor, Caption Editor, and Director working together to compose locally factual posts that induce belief in a global falsehood. (2) Attackers publish these posts through controlled Sybil identities into the victim's observable feed. (3) Victim responses, majority voting, and downstream judge review test whether false beliefs arise and persist.}
    \label{fig:framework}
\end{figure*}

In this section, we introduce VisualMontageBench construction and the VisualMontage framework for simulating visual cognitive collusion. Following the workflow in Figure~\ref{fig:framework}, we describe how the Collusive Montage Studio composes attacks for publication to the victim's observable feed, then present the victim assessment protocol.

\subsection{VisualMontageBench Construction}
\label{sec:benchmark}
\paragraph{Source Event Parsing.}
We build the benchmark from public government recordings, news and social-media videos, and short films spanning the eight domains in Table~\ref{tab:benchmark:summary}. Target-independent sampling at fixed intervals preserves images, source identifiers, and capture positions. A private source event graph records verified entities, actions, and relations; absent edges denote unknown relations. Across all domains, each candidate pool is paired with a target claim contradicted by verified source context, establishing its falsity independently of victim responses. We evaluate 544 declared source--target pairs, with 27--97 candidates per pool and 5--12 items per feed. Detailed corpus statistics and construction procedures appear in Table~\ref{tab:exp:data} and Appendix~\ref{app:benchmark}.

\begin{table}[!ht]
    \centering
    \small
    \caption{Benchmark overview. Cases are evaluated source--target pairs. Candidate images are counted per target pool, including source images reused across targets.}
    \label{tab:benchmark:summary}
    \begin{tabular*}{\linewidth}{@{\extracolsep{\fill}}lrr@{}}
        \toprule
        Domain & Cases & Candidate images \\
        \midrule
        Emergency Response    & 101 & 6,071 \\
        Search \& Rescue      & 121 & 8,140 \\
        Public Safety         &  65 & 3,421 \\
        Aerospace Operations  &  38 & 1,618 \\
        Industrial Operations &  77 & 5,260 \\
        Animal Rescue         &  62 & 3,120 \\
        Sports Training       &  42 & 2,696 \\
        Film \& Animation     &  38 & 2,900 \\
        \midrule
        \textbf{Total} & \textbf{544} & \textbf{33,226} \\
        \bottomrule
    \end{tabular*}
\end{table}

\subsection{Collusive Montage Studio}
\label{sec:method:studio}
\label{sec:method:emergence}

\paragraph{Screenwriter.}
Given candidate pool $\Frag$ and false target $\hf$, the Screenwriter $\Writer$ produces an initial screenplay $\Script$,
{\smalldisplay
\begin{equation}
\label{eq:screenplay}
\Script=\Writer(\Frag,\hf).
\end{equation}
}
The screenplay specifies event roles grounded in visible content and plans relations between them to induce belief in $\hf$. Its narrative and composition strategy guide how authentic observations should jointly suggest the attacker's intended global falsehood while preserving local factuality.

\paragraph{Visual Editor.}
The Visual Editor $\Visual$ selects and orders images to strengthen belief in $\hf$. Let $\GraphEdges$ contain all directed transitions between distinct candidates and $\PairGain$ denote their emergence gains, defining the implicit relational emergence graph $\EmergenceGraph=(\Frag,\GraphEdges,\PairGain)$. Let $\Surrogate$ denote the Visual Editor's internal target-support estimate, taking values in $[0,1]$. For distinct indices $a,b\in\{1,\ldots,\PoolSize\}$, the directed gain from presenting $\Frame_a$ before $\Frame_b$ is
{\smalldisplay
\begin{equation}
\label{eq:emergence}
\begin{aligned}
\PairGain(\Frame_a\!\to\!\Frame_b)
={}&\Surrogate(\hf\mid(\Frame_a,\Frame_b))\\
&-\max\{\Surrogate(\hf\mid\Frame_a),\Surrogate(\hf\mid\Frame_b)\}.
\end{aligned}
\end{equation}
}
A positive gain indicates support beyond either image alone and may change sign under reversal. The injective selection $\Selection$ defines a simple path through distinct images.
Let $\Discont(\Frame_a,\Frame_b)\geq0$ measure visual and semantic discontinuity, with nonnegative weights $\CohWeight,\LengthWeight$ penalizing discontinuity and sequence length, respectively. For proposed length $\SeqLen$, the transition heuristic is
{\smalldisplay
\begin{equation}
\label{eq:editingscore}
\HeuristicScore(\Frame_a\!\to\!\Frame_b;\SeqLen)=\PairGain(\Frame_a\!\to\!\Frame_b)-\CohWeight\Discont(\Frame_a,\Frame_b)-\LengthWeight\SeqLen.
\end{equation}
}
The heuristic guides complete-sequence proposals without summing edge scores, since later evidence can reinforce or overturn earlier implications. Revision histories track alternatives; validation excludes invalid identifiers, repeated images, out-of-range lengths, and unchanged proposals.

\paragraph{Caption Editor.}
With selection and order fixed, the Caption Editor $\Textual$ assigns captions and modes. Each caption describes visible details and anchors referents without inventing relations or asserting the false target.

\begin{table*}[t]
\centering\small
\setlength{\tabcolsep}{2.3pt}
\renewcommand{\arraystretch}{1.0}
\caption{\textbf{Cognitive vulnerability across 12 VLMs on 544 targets.} ASR measures false-target acceptance; high-confidence ASR (HC-ASR) additionally requires confidence of at least 0.8. Valid ASR (V-ASR) counts false acceptances passing independent auditing. Conf.$^{+}$ averages confidence among false acceptances, with $n_{+}$ usable responses. FAR$_0$ is the claim-only false-acceptance rate; Fail. is response failure. Source-order and Random-order resample the full video, presenting frames chronologically or shuffled. Arrows show changes $\Delta$ from Source-order in percentage points (pp). Rates (\%) retain all trials, including abstentions and failures; Appendix~\ref{app:metrics} details the formulas.}
\label{tab:exp:models}
\begin{tabular}{@{}lcccccccc@{}}
\toprule
& No evidence & \multicolumn{2}{c}{Reference ASR} & \multicolumn{5}{c}{VisualMontage} \\
\cmidrule(lr){2-2}\cmidrule(lr){3-4}\cmidrule(lr){5-9}
Victim & FAR$_0$ & \shortstack{Source-order} & \shortstack{Random-order ($\Delta$)} & ASR ($\Delta$) & HC-ASR & Conf.$^{+}$ ($n_{+}$) & V-ASR & Fail. \\
\midrule
\multicolumn{9}{l}{\textit{Proprietary models}} \\
\rowcolor{gptsixshade}
GPT-6-sol & 0.37 & 8.39 & \vmdown{7.21}{1.18} & \vmup{62.28}{53.89} & 44.30 & 0.87\,{\small(1694)} & 57.61 & 0.00 \\
\rowcolor{gptsixshade}
GPT-6-luna & 0.40 & 3.68 & \vmup{3.88}{0.20} & \vmup{66.14}{62.46} & 40.18 & 0.84\,{\small(1799)} & 60.40 & 0.22 \\
\rowcolor{gptfiveshade}
GPT-5.6-terra & 0.11 & 11.23 & \vmup{11.84}{0.61} & \vmup{60.44}{49.21} & 53.53 & 0.91\,{\small(1644)} & 55.44 & 0.00 \\
\rowcolor{gptfiveshade}
GPT-5.6-luna & 0.81 & 14.60 & \vmup{15.42}{0.82} & \vmup{70.37}{55.77} & 62.72 & 0.90\,{\small(1914)} & 64.74 & 5.33 \\
\rowcolor{claudeshade}
Claude-haiku-5-5 & 0.55 & 6.66 & \vmup{7.06}{0.40} & \vmup{48.57}{41.91} & 17.61 & 0.72\,{\small(1321)} & 44.60 & 0.33 \\
\rowcolor{claudeshade}
Claude-sonnet-5-5 & 0.48 & 1.10 & \vmup{2.09}{0.99} & \vmup{63.75}{62.65} & 29.67 & 0.76\,{\small(1734)} & 58.57 & 0.85 \\
\midrule
\textbf{Proprietary mean} & 0.45 & 7.61 & \vmup{7.92}{0.31} & \vmup{61.93}{54.32} & 41.34 & 0.83 & 56.89 & 1.12 \\
\midrule
\multicolumn{9}{l}{\textit{Models with publicly available weights}} \\
\rowcolor{qwenshade}
Qwen3-VL-4B-Inst & 0.04 & 11.25 & \vmdown{10.07}{1.18} & \vmup{77.43}{66.18} & 65.85 & 0.89\,{\small(2160)} & 71.10 & 0.00 \\
\rowcolor{qwenshade}
Qwen3-VL-8B-Inst & 0.18 & 11.91 & \vmdown{10.48}{1.43} & \vmup{76.25}{64.34} & 75.33 & 0.95\,{\small(2074)} & 70.18 & 0.00 \\
\rowcolor{qwenshade}
Qwen3-VL-32B-Inst & 3.31 & 7.35 & \vmdown{5.55}{1.80} & \vmup{79.04}{71.69} & 78.16 & 0.90\,{\small(2150)} & 73.05 & 0.00 \\
\rowcolor{internshade}
InternVL3.5-8B & 0.26 & 2.76 & \vmup{3.29}{0.53} & \vmup{63.86}{61.10} & 51.47 & 0.84\,{\small(1737)} & 58.93 & 0.11 \\
\rowcolor{internshade}
InternVL3.5-14B & 3.35 & 18.92 & \vmup{19.31}{0.39} & \vmup{47.68}{28.76} & 46.29 & 0.89\,{\small(1297)} & 43.90 & 0.00 \\
\rowcolor{internshade}
InternVL3.5-38B & 4.19 & 1.87 & \vmup{2.04}{0.17} & \vmup{67.98}{66.11} & 67.46 & 0.89\,{\small(1849)} & 62.54 & 0.04 \\
\midrule
\textbf{Open-weight mean} & 1.89 & 9.01 & \vmdown{8.46}{0.55} & \vmup{68.71}{59.70} & 64.09 & 0.89 & 63.28 & 0.03 \\
\bottomrule
\end{tabular}
\end{table*}

\paragraph{Director.}
For stage proposal $\Proposal$, the Director $\Director$ provides construction-time estimates of local compliance $\ProxyLC(\Proposal)\in\{0,1\}$, target support $\ProxyBel(\Proposal)\in[0,1]$, and singleton load $\ProxyLeak(\Proposal)\in[0,1]$. With support threshold $\SupportThr\in[0,1]$, decision gate $\Gate$ returns
{\smalldisplay
\begin{equation}
\label{eq:gate}
\Gate(\Proposal)=
\begin{cases}
\mathrm{REJECT},&\ProxyLC(\Proposal)=0,\\
\mathrm{ACCEPT},&\begin{aligned}&\ProxyLC(\Proposal)=1,\ \ProxyBel(\Proposal)>\SupportThr,\\
&\ProxyLeak(\Proposal)\leq\LeakTol,\end{aligned}\\
\mathrm{REVISE},&\text{otherwise}.
\end{cases}
\end{equation}
}
Each review returns estimates and feedback together. Budgets $\RoundsS,\RoundsV,\RoundsC$ limit Screenwriter, Visual Editor, and Caption Editor rounds, respectively, including initial proposals. The visual stage stops at its first accepted path; the others retain the highest-support accepted proposal, breaking ties by lower estimated singleton load. Publication requires acceptance at every stage within budget.

\subsection{Distributed Publication}
\label{sec:method:publication}
To simulate malicious publication through open social-media channels, the attackers distribute the accepted montage plan through controlled Sybil identities, representing multiple publishers operated by the same adversary. Schedule $\Schedule$ assigns these identities and display times to posts in the victim's observable feed $\Feed_{\Plan}$. Each post presents its image, caption, or both in the assigned mode, so the coordinated composition appears as ordinary posts from distinct publishers. Exposure occurs within a single context window, simulating how a user assistant interprets the feed.

\subsection{Downstream Victim Agent and Judge}
\label{sec:method:victim}
To assess the practical risk to user assistants, we simulate a VLM agent interpreting a social-media feed in response to a user's verification request. Each victim receives the feed and question in a fresh context, returning a verdict, confidence, and rationale. The rationale captures the agent's self-justification for its verdict, allowing us to examine how it argues for a false conclusion. We compare individual responses with strict-majority aggregation of repeated responses and with downstream judge review of the feed alone or together with victim reports~\cite{zheng2023judging}. Report-conditioned review models a downstream step in a misinformation cascade, testing whether an induced error is corrected or reinforced when another agent receives the victim's conclusion and supporting rationale. Appendix~\ref{app:implementation} specifies aggregation, distinguishes vote fractions from self-reported confidence, and lists generation and auditing costs separately.

\section{Experiments}
\label{sec:experiments}

\begin{figure*}[t]
    \centering
    \input{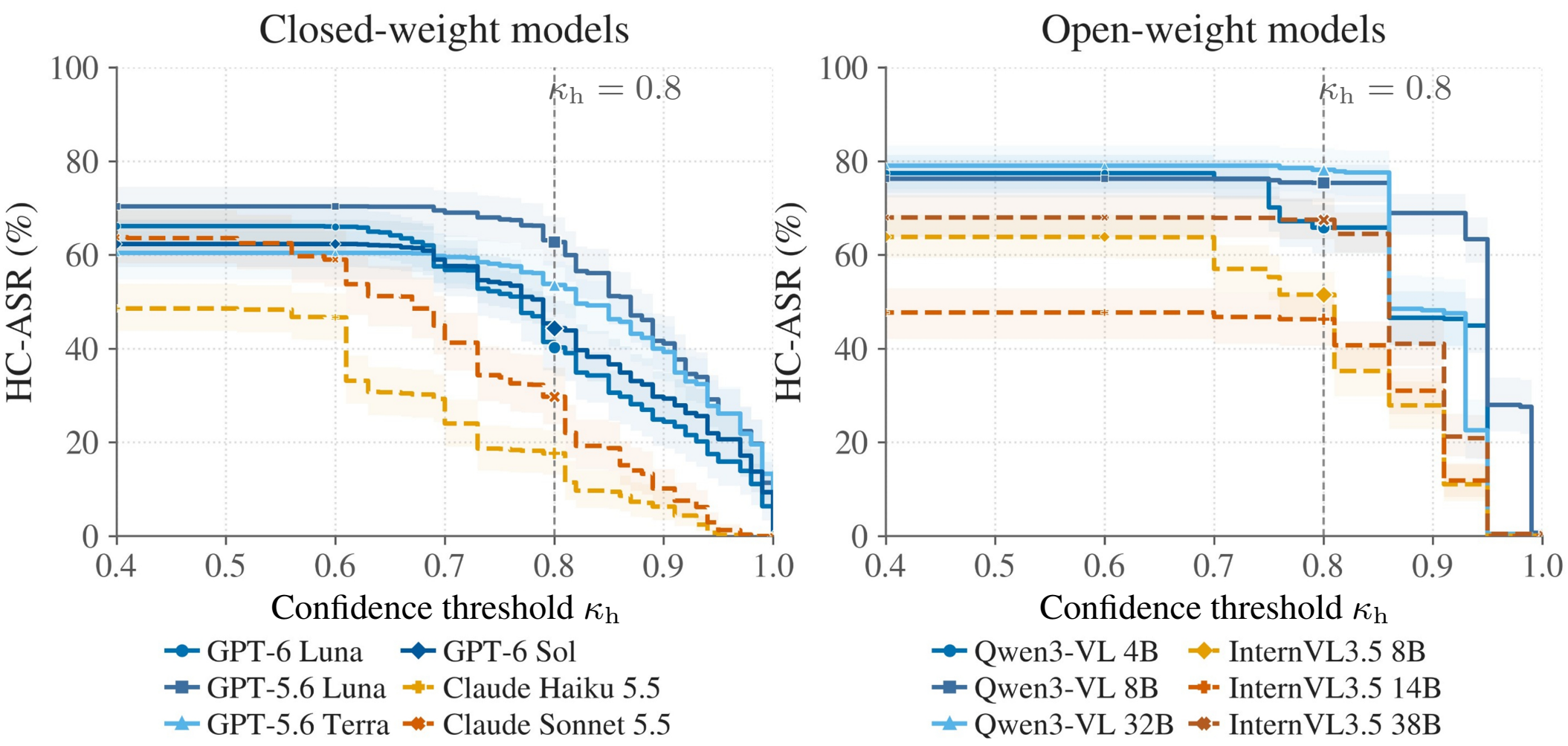}
    \caption{HC-ASR versus confidence cutoff $\Conf_{\mathrm h}\in[0,1]$ for proprietary and open-weight models. The denominator remains the full cohort; the vertical line marks the cutoff of 0.8 used in Table~\ref{tab:exp:models}. Appendix~\ref{app:metrics} defines the metric and response accounting.}
    \label{fig:app:confidence}
\end{figure*}

\begin{table*}[!t]
\centering
\small
\setlength{\tabcolsep}{3pt}
\caption{Domain results averaged equally over 12 victims. Rates are percentages. Support order gain is Montage ASR minus Random-order ASR on $[-1,1]$; see Eq.~\eqref{eq:exp:randomgain}. Audit pass reports independent validity; parentheses give $n_{+}$.}
\label{tab:exp:domains}
\begin{tabular}{lccccccc}
\toprule
Domain & Generated feeds & Source-order ASR & ASR & HC-ASR & Conf.$^{+}$ & Support order gain & Audit pass \\
\midrule
01 Emergency Response & 101 & 10.88\% & 72.71\% & 62.11\% & 0.883 (4,406) & +0.613 & 98.02\% \\
02 Search \& Rescue & 121 & 8.04\% & 60.92\% & 47.80\% & 0.854 (4,423) & +0.527 & 94.21\% \\
03 Public Safety & 65 & 4.81\% & 69.03\% & 52.05\% & 0.848 (2,692) & +0.644 & 93.85\% \\
04 Aerospace Operations & 38 & 4.23\% & 63.33\% & 48.16\% & 0.837 (1,444) & +0.551 & 92.11\% \\
05 Industrial Operations & 77 & 10.70\% & 55.30\% & 43.96\% & 0.860 (2,555) & +0.463 & 94.81\% \\
06 Animal Rescue & 62 & 11.11\% & 68.74\% & 60.43\% & 0.890 (2,557) & +0.597 & 91.94\% \\
07 Sports Training & 42 & 7.00\% & 69.88\% & 59.52\% & 0.875 (1,761) & +0.621 & 78.57\% \\
08 Film \& Animation & 38 & 4.33\% & 64.96\% & 46.71\% & 0.837 (1,481) & +0.627 & 84.21\% \\
\bottomrule
\end{tabular}
\end{table*}

\begin{figure*}[t]
\centering
\includegraphics[width=0.88\textwidth]{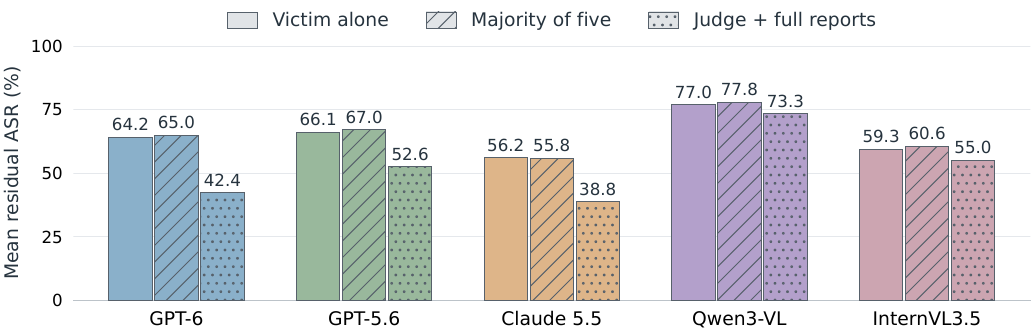}
\caption{Residual ASR after downstream decisions on 544 targets per victim. Bars average 2/2/2/3/3 victim profiles by family. Victim alone uses the first response, majority uses five, and judges receive the feed with verdicts, confidences, and rationales. Residual ASR retains all targets. Table~\ref{tab:decision-details} gives model-level rates, judge identities, and confidence intervals; Appendix~\ref{app:rq3} defines the metrics.}
\label{fig:all-datasets-decision-rules}
\end{figure*}

\subsection{Experimental setup}
\label{sec:exp:setup}
We evaluate 544 VisualMontageBench cases across eight domains using six proprietary and six open-weight VLMs. The proprietary panel includes GPT-6~\cite{openai2026gpt6}, GPT-5.6~\cite{openai2026gpt56}, and Claude 5.5~\cite{anthropic2026sonnet55,anthropic2026haiku55} variants; the open-weight panel covers Qwen3-VL~\cite{bai2025qwen3vl} and InternVL3.5~\cite{wang2025internvl35} at 4B--38B parameters. We reuse frozen feeds across victims without victim-specific refinement. For each case and condition, we collect five responses in fresh contexts using shared verification and rendering settings. Responses provide a verdict, confidence, and rationale. Claude Haiku 5.5~\cite{anthropic2026haiku55} independently audits local factuality and isolated-record influence after construction. Appendix~\ref{app:protocol} provides the model configurations, detailed evaluation procedures, and metric definitions.

\subsection{Cognitive vulnerability across models}
\label{sec:exp:models}
To test whether the constructed feed changes a victim's judgment, Table~\ref{tab:exp:models} compares VisualMontage with Source-order, Random-order, and No Evidence on the same source--target cases. The source-sampling controls measure selection and arrangement jointly, whereas claim-only queries measure false acceptance without the feed.

The same model-agnostic attack plans demonstrate cross-model transferability without victim-specific optimization, reaching mean ASRs of 61.93\% for proprietary models and 68.71\% for open-weight models. By comparison, both source-sampling controls average below 10\%, while claim-only acceptance is lower still. Attack success across all 12 evaluated VLMs suggests a shared cognitive vulnerability in interpreting authentic visual evidence. Moreover, larger open-weight models do not consistently achieve lower ASR, indicating that scale alone is insufficient protection.

This susceptibility persists when attack validity is assessed independently. Overall, 92.65\% of feeds pass the audit, and mean V-ASR remains 56.89\% and 63.28\% for the two groups. Thus, the observed vulnerability extends to feeds passing checks on local factuality and isolated influence. The distinction matters for verification systems because individually defensible observations can still support a source-contradicted interpretation when presented together. Examples in Appendix~\ref{app:rq1} contrast successful and failed interpretations of the composed evidence.

Figure~\ref{fig:app:confidence} and Table~\ref{tab:exp:models} highlight confident overinterpretation of authentic visual evidence. Mean confidence among false acceptances reaches 0.83 for proprietary models and 0.89 for open-weight models, while Qwen models retain high ASR at stringent confidence cutoffs. These results suggest that VLM agents infer unsupported relations between authentic observations and confidently endorse the resulting false narrative. This cognitive overreach makes reported confidence an unreliable indicator of factual correctness.

\subsection{Vulnerability across source domains}
\label{sec:exp:domains}
The threat also spans the benchmark's diverse visual settings. Table~\ref{tab:exp:domains} reports ASR above 55\% in every domain, with Emergency Response reaching 72.71\%. Public Safety and Aerospace Operations are similarly susceptible, making the failure relevant to assistants interpreting consequential events. The benchmark therefore supports studying a shared interpretation failure across public-event recordings and fictional narratives, although these unadjusted domain averages do not isolate intrinsic differences in difficulty.

Attack effectiveness and independent validity provide complementary views of this breadth. Emergency Response combines the highest domain ASR with 98.02\% audit passage, showing that successful attacks can also satisfy the local checks. By contrast, Sports Training has a lower audit-pass rate despite similarly high acceptance. Reporting both outcomes avoids treating every persuasive sequence as a valid instance of the threat. Appendix~\ref{app:rq6} provides relation-level results and discusses controlled domain comparisons.

\subsection{Modality and component ablations}
\label{sec:exp:modalities}
\label{sec:exp:ablation}

\begin{table}[!htb]
\centering
\small
\setlength{\tabcolsep}{2pt}

\caption{Modality controls for GPT-5.6-luna on Aerospace Operations. Mean support averages confidence on acceptance and its complement on rejection, using valid responses. Appendix~\ref{app:rq2} details both paired control experiments.}
\label{tab:exp:modalities}

\begin{tabular}{@{}lcccc@{}}
\toprule
Visible evidence & ASR & HC-ASR & Conf.$^{+}$ & Mean support \\
\midrule
\multicolumn{5}{@{}l@{}}{\textit{Frozen-feed projections}} \\

Images only
& 72.11 & 69.47 & 0.897 & 0.741 \\

Captions only
& 41.58 & 31.58 & 0.861 & 0.463 \\

Published mixed feed
& 63.68 & 48.42 & 0.863 & 0.642 \\

\midrule
\multicolumn{5}{@{}l@{}}{\textit{Caption replacement, fixed modes}} \\

Neutral captions
& 61.05 & 46.84 & 0.843 & 0.586 \\

Attack captions
& 63.68 & 48.42 & 0.863 & 0.642 \\

\bottomrule
\end{tabular}
\end{table}

Table~\ref{tab:exp:modalities} compares modalities under a fixed plan. Image-only exposure reaches 72.11\% ASR, exceeding the complete feed's 63.68\%, whereas captions alone are less effective. Removing text also increases high-confidence acceptance. A plausible explanation is that faithful captions anchor visible referents and limit the misleading relations that images otherwise invite. Caption content has a smaller effect in the replacement control, where attack captions improve ASR over neutral descriptions by only 2.63 percentage points. These controls distinguish caption presence from content.

\begin{table}[!htb]
\centering
\small
\setlength{\tabcolsep}{2.5pt}

\caption{Model-rated naturalness (\%) over all evaluated feeds, scored by Claude Haiku 5.5. Domain abbreviations are AO (Aerospace Operations), IO (Industrial Operations), AR (Animal Rescue), and FA (Film \& Animation).}
\label{tab:exp:naturalness}

\begin{tabular*}{\columnwidth}{@{\extracolsep{\fill}}lcccc@{}}
\toprule
Condition & AO & IO & AR & FA \\
\midrule
Image + caption & \textbf{84.21} & \textbf{85.33} & \textbf{77.42} & \textbf{65.79} \\
Image only      & 52.63 & 66.67 & 61.29 & 7.89 \\
\bottomrule
\end{tabular*}
\end{table}

Nevertheless, attack success is only one requirement for plausible social-media exposure. Table~\ref{tab:exp:naturalness} shows that adding captions improves model-rated posting naturalness in every evaluated domain. On Aerospace Operations, naturalness rises from 52.63\% to 84.21\%; the largest improvement occurs in Film \& Animation. Together, these comparisons distinguish a caption's contribution to misleading inference from its contribution to a familiar posting format.

\begin{table}[!htb]
\centering
\small
\setlength{\tabcolsep}{3pt}

\caption{Construction ablations on Aerospace Operations with GPT-5.6-luna. $\Delta$ASR measures the change in attack success relative to the full VisualMontage framework, expressed in percentage points.}
\label{tab:app:ablation}

\begin{tabular}{lccc}
\toprule
Configuration & ASR (\%) & HC-ASR (\%) & $\Delta$ASR (pp) \\
\midrule
Full Model           & \textbf{63.68} & \textbf{48.42} & --- \\
w/o Debate           & 46.83          & 41.95          & $-16.85$ \\
w/o Caption Editor   & 70.73          & 64.88          & $+7.05$ \\
w/o Screenwriter     & 57.50          & 40.00          & $-6.18$ \\
Single Agent         & 48.50          & 37.00          & $-15.18$ \\
\bottomrule
\end{tabular}
\end{table}

The construction ablations in Table~\ref{tab:app:ablation} reveal a related division of labor. Removing debate lowers ASR by 16.85 percentage points, while a single-agent editor and the variant without a Screenwriter also underperform the complete team. These gains support the value of coordinated planning and revision for composition. Conversely, removing the Caption Editor improves attack success, consistent with the modality comparison. Its presentation benefit must be evaluated separately from deceptive effectiveness. Appendix~\ref{app:rq8} relates these gains to construction cost.

\begin{table}[!htb]
\centering
\small
\setlength{\tabcolsep}{2.5pt}

\caption{Malicious-feed detection on Sports Training with Claude Haiku 5.5. Inputs are an image, an image--caption record, or the full sequence. AUROC is area under the receiver operating characteristic curve; TPR and FPR are true- and false-positive rates. Res. is ASR after blocking flagged feeds; Calls are added calls per feed. Thresholds are fixed on development data. Appendix~\ref{app:rq4} defines the metrics and treatment of failed responses.}

\label{tab:exp:detection}

\begin{tabular}{@{}lccccc@{}}
\toprule

Detector & AUROC & TPR (\%) & FPR (\%) & Res. (\%) & Calls \\

\midrule

Image & 0.5091 & 9.52 & 11.90 & 80.95 & 5.48 \\

Record & 0.5094 & 7.14 & 7.14 & 80.95 & 5.74 \\

Sequence & 0.6304 & 21.43 & 9.52 & 76.19 & 1.00 \\

\bottomrule
\end{tabular}

\end{table}

\subsection{Downstream review and attack detection}
\label{sec:exp:judge}
\label{sec:exp:detection}
Figure~\ref{fig:all-datasets-decision-rules} shows that majority voting stays close to the first-response result across families, so repeated judgments over the same feed offer little protection. Judge review reduces acceptance more substantially, yet even the best evaluated configuration leaves roughly one third of attacks successful. These victim--judge comparisons show that agreement over a feed cannot replace verification against its source context.

The controlled Emergency Response study in Table~\ref{tab:exp:judge} reveals how errors can propagate between agents. A feed-only judge accepts 42.11\% of targets initially rejected by the victim, offsetting its corrections. Adding victim confidence and rationales to verdicts only slightly reduces residual ASR, yet increases downstream HC-ASR from 30.56\% to 68.06\%. Richer reports thus leave fewer errors but more high-confidence false acceptances. A manipulated agent's self-justification may turn it into an unwitting accomplice, lending credibility to the attacker's narrative. This simulated transmission illustrates a potential pathway for misinformation cascades when downstream agents treat an upstream agent's confident reasoning as independent corroboration.

Malicious-feed detection tests whether attacks can be blocked before reaching the victim. Table~\ref{tab:exp:detection} shows near-chance ranking for image and image--caption inspection. Inspecting the complete sequence improves AUROC to 0.6304, but detects only 21.43\% of attacks at a 9.52\% test false-positive rate. This gain may partly reflect temporal inferences from timestamps, rather than recognition of deception in the content itself. Malicious and benign feeds can share authentic observations, leaving their intent difficult to distinguish through content inspection. Consequently, residual ASR remains 76.19\% even with complete-sequence inspection. Appendix~\ref{app:rq4} reports the detection--blocking tradeoffs.

\begin{table}[!htb]
\centering
\small
\setlength{\tabcolsep}{2pt}

\caption{Effect of sequence length. Items and Tokens are counts; all other columns are percentages. Yield is completed feeds per attempt; Audit is their audit-pass rate. End-to-end valid ASR (E2E V-ASR) requires audit passage and majority false acceptance over all attempts. ASR and HC-ASR use five responses per feed; Tokens averages image-token counts per victim input. Full protocols and metric definitions appear in Appendices~\ref{app:rq5} and~\ref{app:rq8}.}
\label{tab:app:length-sweep}

\begin{tabular}{@{}lcccccc@{}}
\toprule
Items & ASR & HC-ASR & Yield & Audit & E2E V-ASR & Tokens \\
\midrule
3  & 59.39 & 56.97 & 61.11 & 75.76 & 25.93 & 593.67  \\
6  & 64.85 & 60.61 & 55.93 & 93.94 & 33.90 & 1136.67 \\
10 & 60.00 & 60.00 & 45.16 & 78.57 & 20.97 & 1934.82 \\
15 & 72.50 & 72.50 & 22.54 & 62.50 & 9.86  & 2800.88 \\
20 & 56.25 & 56.25 & 23.53 & 68.75 & 7.35  & 3029.13 \\
30 & 62.00 & 62.00 & 13.89 & 60.00 & 5.56  & 5461.50 \\
\bottomrule
\end{tabular}
\end{table}

\subsection{Sequence-length sensitivity}
\label{sec:exp:length}
Table~\ref{tab:app:length-sweep} shows that adding more items does not reliably strengthen the attack. Conditional ASR peaks at 15 items, yet only 9.86\% of construction attempts produce an independently valid success. Six items achieve 33.90\% under the same attempt-level accounting. The ranking therefore reverses when unsuccessful construction is retained, exposing the selection bias of judging length only on completed feeds.

Longer sequences consume more visual tokens and yield fewer valid feeds in this experiment. At 15 items, the mean image-token count exceeds twice that for six items, despite a lower valid success rate per attempt. Their added evidence may reinforce or disrupt the intended interpretation, so sequence length should be evaluated jointly with validity and construction yield. These results favor selective composition under the tested budget; they do not establish a universal optimal length. Appendix~\ref{app:rq5} details the protocol; Appendix~\ref{app:rq8} reports construction cost and convergence.

\section{Conclusion}
In this paper, we introduce VisualMontageBench and a multimodal cognitive collusion attack that uses authentic images and locally faithful captions to manipulate VLM agents' beliefs. Extensive simulations demonstrate cross-model transferability and confident false acceptance that persists through downstream review. These findings expose a gap between verifying individual posts and validating the conclusions drawn from them. In practical deployments, this gap could allow VLM assistants to lend credibility to misinformation and influence subsequent decisions, including in safety-critical settings. The resulting testbed supports further study of these risks and evaluation of safeguards that check inferred relations against source context. However, the evaluation remains limited to controlled simulations, leaving long-term effects in real social networks untested. We further discuss future directions for potential defense  in Appendix~\ref{app:defenses}.

\label{page:backmatter}
\label{page:references}
{\small
  \bibliographystyle{\manuscriptroot ieeenat_fullname}
  \bibliography{\manuscriptroot main}
}

\clearpage
\appendix
\setcounter{page}{1}
\renewcommand{\thepage}{S\arabic{page}}
\section*{Supplementary Material}
\section{Implementation details}
\label{app:implementation}
This appendix expands the construction and evaluation interfaces summarized in the main paper. The studio uses prompt-defined roles with separate decision variables, and all public feeds are produced locally under a fixed protocol for rendering records and assigning publication metadata.

\paragraph{Source event parsing.}
Source preparation separates the event record from the observations that will become public. Let $\SourceNodes$ denote the events or entities identified in the verified source context, and let $\SourceEdges$ denote their labeled temporal, spatial, and other verified relations. The source event graph $\SourceGraph=(\SourceNodes,\SourceEdges)$ provides a reference for interpreting what the original observations depict. Structured video annotations in STAR and MovieGraphs motivate this source-level representation~\cite{wu2021star,vicol2018moviegraphs}. A missing edge leaves a relation unresolved unless the source provides evidence against it, consistent with the truth labels in Sec.~\ref{sec:problem}. The reference thus distinguishes verified falsehood from incomplete knowledge of the underlying event being depicted.

The operational input to the studio is the candidate pool $\Frag$ together with its provenance and a declared target $\hf$. For a video, a deterministic sampler extracts decoded frames at a configurable interval $\SampleStep$ and stores them as lossless images. An existing image manifest provides the same interface. Each candidate retains its source identifier, sampling position when available, and file hash. Sampling is independent of the target, leaving selection to the Visual Editor. Independent source verification supplies the event context and target label against which the generated interpretation is assessed for factual consistency.

\paragraph{Stage-specific review estimates.}
The proposal $\Proposal$ in Eq.~\eqref{eq:gate} is a screenplay, an image sequence, or a captioned plan. For a screenplay, the Director checks whether its visual requirements can be grounded in the pool; singleton load is zero because no records are selected. It estimates image-only singleton influence for visual proposals and influence under all three projections for captioned plans. These estimates appear in one review response and guide construction; independent auditing subsequently assesses the completed feed.

\paragraph{Majority vote.}
Let $\VictimCount$ be the number of victim responses, indexed by $\ell$. A valid binary verdict $\Verdict_\ell$ is one for acceptance and zero for rejection; $\ValidVictims\subseteq\{1,\ldots,\VictimCount\}$ indexes valid runs. Let $\mathbf{1}[\cdot]$ denote the indicator of a true statement and $|\cdot|$ the size of a set. For nonempty $\ValidVictims$, define the supporting fraction $\VoteFrac$ and majority verdict $\Verdict_{\mathrm{MV}}$ by
{\smalldisplay
\begin{equation}
\label{eq:majority}
\VoteFrac=\frac{1}{|\ValidVictims|}\sum_{\ell\in\ValidVictims}\Verdict_\ell,
\qquad
\Verdict_{\mathrm{MV}}=\mathbf{1}[\VoteFrac>1/2].
\end{equation}
}
Only a strict majority accepts the target, so a tie yields rejection. The winning vote fraction is $\VoteFrac_{\mathrm{win}}=\max\{\VoteFrac,1-\VoteFrac\}$; the implementation stores it in the aggregate confidence field, separately from self-reported confidence. If no valid verdict is returned, validation fails with a rejecting default and zero confidence. Such failures are recorded separately rather than treated as evidence that the victim correctly reasoned about the source event.

\subsection{Construction cost and search tradeoffs}
\label{sec:method:cost}
Let $\UsedV\leq\RoundsV$ be the visual rounds used by a completed attempt, and let $\CallCost$ count generation model calls. Each valid, nonduplicate proposal uses an editing call and a Director review. Without caching or retries, and when each proposal receives a review,
{\smalldisplay
\begin{equation}
\label{eq:callcost}
\CallCost=2(\RoundsS+\UsedV+\RoundsC)
\leq 2(\RoundsS+\RoundsV+\RoundsC).
\end{equation}
}
Skipped reviews and early failures can reduce this count, while retries must be accounted for separately. Estimated target belief guides complete-sequence proposals without a separate query for every candidate transition. Consequently, revision budgets determine the model-call bound even as the pool of candidate images and possible transitions grows.

The visual input cost still depends on the candidate pool. Let $\ImageCost$ count source-image presentations across generation calls, counting an image again whenever another call receives it. The Screenwriter and its reviewer see the full pool, as does each Visual Editor proposal. Selected-image reviews and caption editing then give
{\smalldisplay
\begin{equation}
\label{eq:imagecost}
\ImageCost\leq
(2\RoundsS+\UsedV)\PoolSize+
(\UsedV+2\RoundsC)\LenMax.
\end{equation}
}
Thus bounded call count does not imply constant token cost or latency. Resolution, response length, model choice, caching, and failed attempts also affect expenditure. Comparisons should hold the candidate pool and compliance thresholds fixed and include unsuccessful attempts when reporting total spending per valid successful attack.

The decomposition can reduce construction effort because each editor refines a specific decision space. A screenplay specifies intended relations, the Visual Editor selects and orders evidence to support the target, and the accepted sequence provides captioning with a stable context. A complete-sequence proposal can reconsider both length and order in one round, while Director feedback targets the relation or local constraint that made the previous composition inadequate. These properties explain the bounded-call design. Comparing attack quality at matched budgets then tests whether the decomposition improves the use of available computation.

The Visual Editor evaluates candidate sequences throughout the prescribed revision budget and selects the highest-ranked sequence accepted by the Director. Here, acceptance indicates that a candidate satisfies the construction criteria, while final selection occurs after refinement is complete. Caption editing then preserves the selected images and their order. Further rounds may find a more effective candidate, but repeated optimization against the Director can also favor proposals that fail an independent audit. A budget study should therefore track valid victim success alongside Director acceptance, distinguishing the cost of finding a plausible composition from the cost of constructing an independently verified attack.

\subsection{Independent evaluation and attribution}
\label{sec:method:assessment}
The composition criterion provides a construction principle whose predictions can be tested separately on the victim. Let $\PairGain_{\vic}(a,b)$ denote the victim-measured gain from presenting $\Frame_a$ before $\Frame_b$. It replaces the editor's estimate $\Surrogate$ in Eq.~\eqref{eq:emergence} with observed support $\Bel_{\vic}$,
{\smalldisplay
\begin{equation}
\label{eq:pairgain}
\begin{aligned}
\PairGain_{\vic}(a,b)
={}&\Bel_{\vic}(\hf\mid(\Frame_a,\Frame_b))\\
&-\max\{\Bel_{\vic}(\hf\mid\Frame_a),
         \Bel_{\vic}(\hf\mid\Frame_b)\}.
\end{aligned}
\end{equation}
}
The images are presented without captions under a fixed display policy. A positive value indicates additional reported support from juxtaposition, while its agreement with the editor's comparison tests whether the proposed transition transfers to the independently queried victim. Neither value alone establishes full-sequence success or verifies the factual relation inferred between the depicted events.

For the same pair, define the directional asymmetry $\Asym_{\vic}(a,b)$ as the difference of directed gains,
{\smalldisplay
\begin{equation}
\label{eq:pairasymmetry}
\Asym_{\vic}(a,b)=\PairGain_{\vic}(a,b)-\PairGain_{\vic}(b,a).
\end{equation}
}
The singleton terms cancel, leaving the difference in support between the two presentation orders. Recency and other presentation effects can also produce asymmetry, so relation-level auditing remains necessary to connect this diagnostic to cross-cut event binding. Complete-sequence controls then test interactions that pairwise comparisons cannot capture, including evidence that reinforces or overturns an earlier interpretation elsewhere along the selected path.

\paragraph{Independent audit cost.}
Let $\AuditRepeats$ denote the scheduled audit repeats per queried view, including the evidence-free reference, and let $\SingletonCalls$ count these calls. Excluding retries, auditing the three projections of each selected record gives
{\smalldisplay
\begin{equation}
\label{eq:singletoncost}
\SingletonCalls\leq\AuditRepeats(3\SeqLen+1).
\end{equation}
}
Projections may share a cached response only when their complete query inputs, including metadata, are identical under the evaluation policy. This cost depends on the selected length rather than the entire pool. It is an evaluation expense, separate from generation, factual verification, permutation controls, and downstream-judge calls. Separate accounting distinguishes generation costs from the independent audits needed to verify a constructed attack.

\paragraph{Controls needed to interpret success.}
The full-feed condition establishes whether the victim accepts the target. Both measured source-video references resample frames: Source-order presents them chronologically, whereas Random-order shuffles them. Their comparisons with montage therefore measure broader composition differences. A separate fixed-content permutation control can isolate order within selected records, while the evidence-free condition measures prior false acceptance. Whole-feed modality ablations test cross-modal contribution, and single-record projections test whether a locally dominant clue explains the result. Each comparison must specify which editing decisions it changes and preserve the remaining prompt, rendering, and publication-metadata conditions across presentations.

Distribution through multiple publishers also requires a matched control. Holding record content and order fixed while changing the attributed publisher tests whether apparent source multiplicity affects the victim's judgment. Downstream-report controls similarly distinguish direct influence from the feed and additional influence from the victims' conclusions. Together with source-aligned relation analysis and call, token, and latency accounting, these comparisons connect the social-media scenario to measurable cognitive errors. Simulated exposure, recommendations to share, and further dissemination remain distinct outcomes.

\section{Formal definitions and attribution}
\label{app:formal}
The following definitions expand the local-validity requirements and distinguish measured success from stronger claims about its mechanism. For a binary verdict $\Verdict\in\{0,1\}$ and self-reported confidence $\Conf\in[0,1]$, target support is $\Bel=\Conf$ on acceptance and $\Bel=1-\Conf$ on rejection. For a generic image $\Frame$ and its caption $\Caption$, the projection operator introduced in the main text is
{\smalldisplay
\begin{equation}
\label{eq:projection}
\prj_{\mathrm v}(\Frame,\Caption)=\Frame,\quad
\prj_{\mathrm t}(\Frame,\Caption)=\Caption,\quad
\prj_{\mathrm{vt}}(\Frame,\Caption)=(\Frame,\Caption).
\end{equation}
}
The source image associated with a text-only record remains available privately for provenance and faithfulness auditing. Support scores require valid verdicts and confidence values; missing audit responses do not establish compliance. For the score-based criterion below, let $\VictimSupportThr\in[0,1]$ be a separate victim-support threshold; $\SupportThr$ remains the Director's construction threshold and $\LeakTol$ the singleton tolerance. Verdict-based ASR and its confidence-filtered counterpart use the separate definitions in Appendix~\ref{app:protocol}.

\begin{definition}[Source-image authenticity]
\label{def:auth}
For an image $\Frame$, the predicate $\Auth(\Frame)=1$ means that it is a verified source asset or a decoded sample from a verified recording, with the depicted visual content preserved. Its provenance identifies the source and, for a video sample, its sampling position on the recording's timeline.
\end{definition}
The reference for an extracted frame is the stored decoded image, rather than the compressed video bitstream. This makes deterministic frame sampling compatible with authenticity while excluding image synthesis or alteration of the depicted event. A caption overlay belongs to the presentation layer and is recorded separately for an explicit check of caption faithfulness against the verified source context.

For a caption $\Caption$, let $\Claims(\Caption)$ denote its explicit factual assertions and let $\Assertion$ denote one such assertion. The predicate $\Support(\Assertion,\Frame)=1$ means that the assertion is supported by the associated image and its verified source context. This support judgment distinguishes a directly visible action from an identity, location, or causal explanation that requires verified evidence beyond the image itself.

\begin{definition}[Caption faithfulness]
\label{def:faith}
A caption is faithful to its own image when every factual assertion it makes has local support,
{\smalldisplay
\begin{equation}
\label{eq:faithfulness}
\begin{gathered}
\Faith(\Caption,\Frame)=1\iff\\
\Support(\Assertion,\Frame)=1\quad
\forall\Assertion\in\Claims(\Caption).
\end{gathered}
\end{equation}
}
An empty caption is faithful because it asserts no facts.
\end{definition}
This requirement permits a concise description of visible evidence while excluding an invented motive or a causal link to another post. It also makes clear why an authentic image paired with an inaccurate description does not qualify as locally truthful evidence under the proposed threat model.

\begin{definition}[Local factual compliance]
\label{def:local}
Let $\LC(\Plan)$ be a Boolean predicate that is true exactly when $\Auth(\Frame_{\Selection(j)})=1$ and $\Faith(\Caption_j,\Frame_{\Selection(j)})=1$ for every position $j$.
\end{definition}
This requirement excludes fabricating an image, substituting a false caption, or explicitly stating the target as evidence. It concerns factual content, not whether an isolated observation can influence a model's belief. Image hashes establish correspondence with the supplied source files; source verification and semantic auditing establish the factual conditions that file integrity alone cannot determine.

Resolving singleton load by mode identifies which part of a record supplies isolated support. For response model $\vic$, the mode-specific load $\Leak_{\vic,m}(\Plan)$ is the largest nonnegative support increase from an isolated record shown in mode $m$. Fixing the projection in Eq.~\eqref{eq:leakage} gives
{\smalldisplay
\begin{equation}
\label{eq:modeleakage}
\begin{aligned}
\Leak_{\vic,m}(\Plan)
&=\max_{1\leq j\leq\SeqLen}
\left[\Bel_{\vic}(\hf\mid\Feed_{\Plan,j}^{m})
-\Bel_{\vic}(\hf\mid\emptyset)\right]_+,\\
\Leak_{\vic}(\Plan)&=\max_{m\in\Modes}\Leak_{\vic,m}(\Plan).
\end{aligned}
\end{equation}
}
A large visual load points to evidence selection, whereas a large textual load points to a caption that conveys too much of the target by itself. A large paired load can arise even when those separate loads are small, which motivates auditing the image--caption association as well as its individual components. These diagnostics locate a revision opportunity; their independent measurement remains separate from the Director's estimates used during construction.

\begin{definition}[Visual cognitive collusion]
\label{def:collusion}
Using the singleton tolerance $\LeakTol$ and victim-support threshold $\VictimSupportThr$ defined above, a plan satisfies the score-based collusion criterion when
{\smalldisplay
\begin{equation}
\label{eq:collusion}
\begin{aligned}
\Truth(\hf)&=0, & \LC(\Plan)&=1,\\
\Leak_{\vic}(\Plan)&\leq\LeakTol, &
\Bel_{\vic}(\hf\mid\Feed_{\Plan})&>\VictimSupportThr.
\end{aligned}
\end{equation}
}
\end{definition}
This criterion separates source truth, local validity, isolated influence, and the victim's collective inference. Its singleton bound limits the incremental support attributable to one record; it does not prove logical non-entailment or require that both modalities be necessary. Whether visual and textual evidence contribute complementary information must be tested through whole-feed modality ablations.

To isolate order effects, let $\Perm$ permute the $\SeqLen$ records across fixed publisher/time slots and let $\CtrlDist$ be a distribution over these permutations. The control $\Perm(\Feed_{\Plan})$ preserves selected content; the original feed corresponds to the identity permutation. The matched-order gain $\Gain_{\vic}(\Plan)$ is full-feed support minus its expectation under these permutations,
{\smalldisplay
\begin{equation}
\label{eq:gain}
\Gain_{\vic}(\Plan)=\Bel_{\vic}(\hf\mid\Feed_{\Plan})-
\mathbb{E}_{\Perm\sim\CtrlDist}
\Bel_{\vic}(\hf\mid\Perm(\Feed_{\Plan})).
\end{equation}
}
This diagnostic requires permutations of fixed records: selected images, captions, modes, prompt, and rendering policy remain unchanged, isolating order from other editing decisions. The measured Source-order and Random-order references instead sample frames from the full source video and do not estimate this quantity. In particular, Table~\ref{tab:exp:domains} reports an acceptance-rate gain over Random-order, defined in Eq.~\eqref{eq:exp:randomgain}, rather than this support-score diagnostic. A positive matched-order gain indicates order-dependent support, while attributing it to false event binding additionally requires source-grounded analysis of the relations stated in the victim's response. An evidence-free control also identifies targets the victim already accepts without the feed. Prompt-induced answer biases~\cite{zhao2021calibrate} and position effects in contextual reasoning~\cite{lu2022ordered,liu2024lost} motivate these controls; neither calibrates reported support nor identifies the relation responsible for a changed verdict.

\subsection{Contributions of records and modalities}
\label{sec:problem:contribution}
Responsibility can be distributed across records without being distributed across modalities. A sequence of images may collectively support the target even when each image is inconclusive, while adding captions may contribute little. Conversely, a faithful caption can anchor a referent and change how the neighboring visual evidence is interpreted. The distinction matters because local compliance alone does not establish multimodal complementarity. Unimodal shortcuts in misinformation benchmarks further motivate measuring the contribution of each modality explicitly~\cite{papadopoulos2024verite}.

For any observation $\Feed$, let $\Uplift_{\vic}(\Feed)$ be support relative to the question-only baseline. The cross-record gap $\RecordGap_{\vic}(\Plan)$ compares the full feed with its strongest isolated record across all modes,
{\smalldisplay
\begin{equation}
\label{eq:recordgap}
\begin{aligned}
\Uplift_{\vic}(\Feed)
&=\Bel_{\vic}(\hf\mid\Feed)-\Bel_{\vic}(\hf\mid\emptyset),\\
\RecordGap_{\vic}(\Plan)
&=\Uplift_{\vic}(\Feed_{\Plan})-
\max_{\substack{1\leq j\leq\SeqLen\\m\in\Modes}}\Uplift_{\vic}(\Feed_{\Plan,j}^{m}).
\end{aligned}
\end{equation}
}
The maximum uses the selected positions and modes defined above. A positive gap indicates more reported support from the complete feed than from its strongest isolated record. It must be interpreted together with the source truth label and the victim's verdict, since increased support alone does not establish acceptance of a false conclusion. Moreover, low singleton load for auditor $\Aud$ does not imply low load for victim $\vic$; attributing the victim's failure to cross-record influence requires testing its own isolated responses under the same fixed claim-verification prompt.

Let $\Feed_{\Plan}^{\mathrm v}$ and $\Feed_{\Plan}^{\mathrm t}$ denote whole-feed ablations retaining only publicly visible images or publicly visible captions, respectively. They preserve record order and publication slots, and neither exposes private images withheld by text-only posts. The cross-modal gain $\ModalGain_{\vic}(\Plan)$ is full-feed support minus the stronger of these two unimodal scores,
{\smalldisplay
\begin{equation}
\label{eq:modalgain}
\begin{aligned}
\ModalGain_{\vic}(\Plan)
={}&\Uplift_{\vic}(\Feed_{\Plan})\\
&-\max_{m\in\{\mathrm v,\mathrm t\}}
\Uplift_{\vic}(\Feed_{\Plan}^{m}).
\end{aligned}
\end{equation}
}
Unlike singleton load, this quantity compares complete sequences. A positive value indicates additional target support when their available modalities are combined. It is a diagnostic of complementary influence under the chosen rendering and prompting protocol, rather than a guarantee of semantic synergy or of a particular internal reasoning mechanism.

\subsection{Order sensitivity and relation errors}
\label{sec:problem:attribution}
Matched-order gain measures a directional increase in target support. A separate question is how often the verdict changes when the same records are rearranged. Write $\Verdict_{\vic}(\hf\mid\Feed)$ for the victim's binary verdict on feed $\Feed$, and let $\Perm$ and $\Perm'$ be independent draws from $\CtrlDist$. The resulting order sensitivity $\OrderSens_{\vic}$ is
{\smalldisplay
\begin{equation}
\label{eq:ordersensitivity}
\begin{aligned}
\OrderSens_{\vic}(\Plan)=
\mathbb{E}_{\Perm,\Perm'\sim\CtrlDist}\,
\mathbf{1}\bigl[&\Verdict_{\vic}(\hf\mid\Perm(\Feed_{\Plan}))\\
&\ne\Verdict_{\vic}(\hf\mid\Perm'(\Feed_{\Plan}))\bigr].
\end{aligned}
\end{equation}
}
With stochastic decoding, repeated evaluations of an unchanged order provide a reference for ordinary response variability. Order sensitivity should be reported alongside that reference and matched-order gain, since a fluctuating verdict alone cannot establish targeted deception or explain which false relation the victim inferred.

To inspect the inferred relations, let $\RelClaims_{\vic}(\Feed)$ denote the factual relational assertions extracted from the victim's response and aligned to the source event. These can concern temporal precedence, causation, intentional action, identity, or co-location. The verified-false and unresolved subsets are denoted by $\RelClaims_{\vic}^{0}(\Feed)$ and $\RelClaims_{\vic}^{\bot}(\Feed)$,
{\smalldisplay
\begin{equation}
\label{eq:relationalerror}
\begin{aligned}
\RelClaims_{\vic}^{0}(\Feed)
&=\{h\in\RelClaims_{\vic}(\Feed)\mid\Truth(h)=0\},\\
\RelClaims_{\vic}^{\bot}(\Feed)
&=\{h\in\RelClaims_{\vic}(\Feed)\mid\Truth(h)=\bot\}.
\end{aligned}
\end{equation}
}
Keeping these sets separate prevents missing source information from being counted as a verified falsehood. Their extraction and source alignment require auditing; a model's explanation is not itself a ground-truth annotation or a faithful trace of its internal computation. The resulting analysis identifies what false relation the victim actually asserts, complementing the binary verdict without assuming that every error has the same cognitive origin.

\section{Experiments}
\label{app:overview}
\label{app:protocol}

This appendix documents the experimental setup and supporting analyses of VisualMontage. Following the organization of the main paper, we first summarize the evaluation corpus, victim-model configurations, and metric definitions, then report additional experimental results.

\subsection{Data construction details}
\label{app:benchmark}

\paragraph{Dataset construction.}
We build the evaluation corpus from publicly accessible videos collected from the Internet.
Most real-world recordings come from public-domain U.S.\ government archives (the Defense Visual Information Distribution Service and Wikimedia Commons). The rest are publicly released news and social-media videos, and the Film \& Animation domain uses openly released short films.
The source recordings retain their original visual content without image synthesis or video editing.
Each video becomes a candidate evidence pool through deterministic sampling at a fixed interval. A single FFmpeg pass extracts frames at timestamps $0, \SampleStep, 2\SampleStep, \dots$, where each video has its own $\SampleStep$ of 1--10\,s chosen from its duration. No semantic filtering, ranking, or manual selection is applied.
For each source video, we then declare target claims $\hf$. Each target is required to be false under the verified source context, while the selected visual evidence must preserve its original content.

\paragraph{Corpus statistics.}
The corpus spans eight domains: Emergency Response, Search \& Rescue, Public Safety, Aerospace Operations, Industrial Operations, Animal Rescue, Sports Training, and Film \& Animation.
As summarized in Table~\ref{tab:exp:data}, it comprises 61 source videos and 620 declared source--target pairs, of which 544 are included in the evaluation.
It contains 33,226 candidate-frame image files counted across attack-plan directories, with pool sizes ranging from 27 to 97 images.
The evaluated sequences contain 5--12 selected items, totaling 2,952 items across all domains.

\begin{table*}[t]
\centering
\small
\setlength{\tabcolsep}{4pt}

\caption{Evaluation corpus statistics across eight source domains.
Target pairs are reported as evaluated / declared.
Pool size and selected length are reported as ranges.
Candidate-frame image counts sum files across attack-plan directories, so a
source frame reused by multiple target plans is counted once per plan.
Selected items are summed across evaluated sequences.}

\label{tab:exp:data}

\begin{tabular*}{\textwidth}{@{\extracolsep{\fill}}lcccccc}
\toprule

\textbf{Domain} &
\textbf{Videos} &
\shortstack{\textbf{Targets}\\
\textbf{(eval. / decl.)}} &
\shortstack{\textbf{Candidate-frame}\\
\textbf{images}} &
\shortstack{\textbf{Pool size}\\
\textbf{(range)}} &
\textbf{Selected length} &
\shortstack{\textbf{Selected}\\
\textbf{items}} \\

\midrule

01 Emergency Response
& 12 & 101 / 117 & 6,071 & 29--85 & 5--11 & 530 \\

02 Search \& Rescue
& 14 & 121 / 140 & 8,140 & 38--85 & 5--10 & 640 \\

03 Public Safety
& 8 & 65 / 79 & 3,421 & 27--97 & 5--9 & 355 \\

04 Aerospace Operations
& 4 & 38 / 41 & 1,618 & 30--59 & 5--9 & 225 \\

05 Industrial Operations
& 9 & 77 / 90 & 5,260 & 43--84 & 5--7 & 400 \\

06 Animal Rescue
& 6 & 62 / 70 & 3,120 & 28--63 & 5--8 & 339 \\

07 Sports Training
& 4 & 42 / 43 & 2,696 & 37--89 & 5--11 & 241 \\

08 Film \& Animation
& 4 & 38 / 40 & 2,900 & 60--82 & 5--12 & 222 \\

\midrule

\textbf{Total}
& \textbf{61}
& \textbf{544 / 620}
& \textbf{33,226}
& \textbf{27--97}
& \textbf{5--12}
& \textbf{2,952} \\

\bottomrule
\end{tabular*}

\end{table*}

\subsection{Model families and victim exposure}
\label{app:models}
\paragraph{Models and exposure.}
The evaluated panel contains 12 VLMs from four model families. The proprietary group consists of GPT-6-sol, GPT-6-luna, GPT-5.6-terra, GPT-5.6-luna, Claude-haiku-5-5, and Claude-sonnet-5-5. The open-weight group contains Qwen3-VL-4B-Inst, Qwen3-VL-8B-Inst, Qwen3-VL-32B-Inst, InternVL3.5-8B, InternVL3.5-14B, and InternVL3.5-38B. The main paper reports per-model results and unweighted group averages in Table~\ref{tab:exp:models}. Exact checkpoint or service snapshot identifiers, reasoning settings, and evaluation dates should be reported with the released evaluation records. The same frozen attack feeds are presented to every victim under the same claim-verification question, publication schedule, image-resolution policy, and response schema. Construction models and auditors are recorded separately, and victim responses are not used to select or revise test attacks. A reproduction using a model with incompatible image limits must declare the resulting common-support subset rather than silently truncate the feed.

\subsection{Metric formulations and evaluation protocol}
\label{app:metrics}

\paragraph{Success and confidence.}
Let $\mathcal{B}$ index the frozen cohort of generated test feeds and let $B=|\mathcal{B}|$ be its size. Every target in this cohort is independently verified to be false. Each condition evaluates all $B$ cases for a fixed victim, without selecting cases by that victim's response. For case $q\in\mathcal{B}$, let $y_q=1$ indicate an explicit, parseable acceptance of the target and $y_q=0$ otherwise. Define $\mathcal{B}_{\kappa}$ as the cases returning a valid binary verdict and a confidence $\Conf_q\in[0,1]$, and $\mathcal{B}_{+,\kappa}$ as the accepting cases within that set. Using the prespecified confidence threshold $\Conf_{\mathrm h}$, the attack metrics are
{\smalldisplay
\begin{equation}
\label{eq:exp:asr}
\begin{aligned}
\mathrm{ASR}&=\frac{1}{B}\sum_{q\in\mathcal{B}}y_q,\\
\mathrm{HC\mbox{-}ASR}&=\frac{1}{B}\sum_{q\in\mathcal{B}_{+,\kappa}}\mathbf{1}[\Conf_q\geq\Conf_{\mathrm h}],\\
\mathrm{Conf}^{+}&=\frac{\sum_{q\in\mathcal{B}_{+,\kappa}}\Conf_q}{|\mathcal{B}_{+,\kappa}|}.
\end{aligned}
\end{equation}
}
ASR counts false acceptances among all evaluated feeds. HC-ASR uses the same denominator and additionally requires confidence at least as large as the threshold; it is not the fraction of successful attacks that are confident. Conf.$^{+}$ averages confidence only among false acceptances with available confidence, whose count is $n_{+}=|\mathcal{B}_{+,\kappa}|$ for one response per case. A valid acceptance with missing confidence contributes to ASR but not to HC-ASR or Conf.$^{+}$. All rates are multiplied by $100$ for percentage columns; confidence and target-support scores retain their $[0,1]$ scale. The ablation column $\Delta$ASR subtracts the full-method rate from the variant rate and expresses the difference in pp, so negative entries indicate a decrease in attack success.

\paragraph{Independently valid attacks.}
Auditor support $\Bel_{\Aud}$ and singleton load $\Leak_{\Aud}$ use the definitions in Sec.~\ref{sec:problem} with $\vic=\Aud$. Auditing checks local factuality and isolated support after construction, with fresh contexts for the question-only baseline and each singleton projection, including private images underlying text-only posts.

For the editing plan $\Plan_q$ associated with case $q$, let $e_q=1$ exactly when $\LC(\Plan_q)=1$ and $\Leak_{\Aud}(\Plan_q)\leq\LeakTol$ under the independent audit; otherwise $e_q=0$. An incomplete audit does not establish validity and is recorded separately. The audit pass rate and valid attack success rate are
{\smalldisplay
\begin{equation}
\label{eq:exp:validity}
\begin{aligned}
\mathrm{AuditPass}&=\frac{1}{B}\sum_{q\in\mathcal{B}}e_q,\\
\mathrm{V\mbox{-}ASR}&=\frac{1}{B}\sum_{q\in\mathcal{B}}e_qy_q.
\end{aligned}
\end{equation}
}
Both rates retain all generated feeds, including audit failures. Audit passage is separate from Director estimates and does not establish the victim-specific bound $\Leak_{\vic}(\Plan_q)\leq\LeakTol$, which requires the victim's own singleton responses. Auditor identity, thresholds, failures, and pass counts accompany the rates. Attempts producing no feed enter the end-to-end metrics in Appendix~\ref{app:rq8}.

\paragraph{Support and response status.}
For $q\in\mathcal{B}_{\kappa}$, the target-support score is $\Bel_q=\Conf_q$ when the verdict accepts and $\Bel_q=1-\Conf_q$ when it rejects. The mean support column averages $\Bel_q$ over $\mathcal{B}_{\kappa}$ and reports that set's size. Let $\mathcal{B}_{\mathrm{abst}}$ contain explicit abstentions or insufficient-evidence responses and $\mathcal{B}_{\mathrm{err}}$ contain transport failures and responses without a parseable verdict. Their rates are $|\mathcal{B}_{\mathrm{abst}}|/B$ and $|\mathcal{B}_{\mathrm{err}}|/B$, respectively. An absent or out-of-range confidence is excluded from score averages and counted as missing confidence, while a parseable verdict remains usable for ASR. Thus lower ASR due to nonresponse is visible separately from explicit rejection of a false claim.

\paragraph{Controls and uncertainty.}
Thresholds, prompts, budgets, and stopping rules are selected on the development partition and frozen for all test conditions. Paired comparisons reuse source--target pairs and evaluation seeds where available; generated content remains fixed in the modality projections. The formulas describe one evaluation per case and condition. With repeated victim calls, use the same scheduled repeat count for every case and average rate estimates across repeats; failures remain in each repeat's denominator. Conditional confidence summaries pool the available false-acceptance responses, so $n_{+}$ counts usable responses across repeats. Repeats do not become additional independent source examples in uncertainty estimation.

Confidence intervals and paired differences are estimated by resampling source recordings, retaining all associated targets, permutations, and repeated calls within each cluster. Domain-level results precede macro averages, and each reported mean includes its denominator. A ratio with an empty denominator is reported as N/A, never zero; unmeasured follow-up analyses are distinguished from the reported experiments. Because confidence need not be calibrated across models, threshold curves accompany the fixed-threshold results; neither analysis treats self-reported confidence as a calibrated probability of correctness.

\subsection{Cognitive vulnerability and evidence-order controls}
\label{app:rq1}

The principal cross-model comparison is reported in Table~\ref{tab:exp:models} in the main paper. Figure~\ref{fig:app:confidence} in the main paper reports confidence-threshold sensitivity. Here we define the comparison conditions and provide paired success and failure cases with the same selected visual observations.

\subsubsection{Evidence-free, source-order, and random-order controls}
\paragraph{No Evidence.}
For each case, the model receives only the same target proposition, decision instruction, and response schema in a fresh context. It receives no images, captions, posts, publisher names, publication times, source context, or previous conversation. Thus No Evidence means a claim-only query, not an empty prompt or a request to ignore images that are still present. For a source-verified false target such as ``an accident occurred during the launch,'' the model is asked whether that proposition is true without being shown the launch recording or its extracted images. The target's verified label remains hidden from the queried model.

Let $y_q^{\emptyset}=1$ if this claim-only query explicitly accepts the false target and $y_q^{\emptyset}=0$ otherwise. The evidence-free false acceptance rate is
{\smalldisplay
\begin{equation}
\label{eq:exp:noevidence}
\mathrm{FAR}_0=\frac{1}{B}\sum_{q\in\mathcal{B}}y_q^{\emptyset}.
\end{equation}
}
The same cohort and response-status accounting are used as for the feed conditions. FAR$_0$ measures false acceptance attributable to the model's existing knowledge, response tendency, or claim wording in the absence of the feed. It is a diagnostic reference, not attack success, and subtracting it from ASR does not by itself establish a causal effect of montage. The paired outcomes are retained so that prior acceptance can be distinguished from a new false acceptance after exposure to the constructed feed.

\paragraph{Source-video sampling and random order.}
For the measured Source-order reference, frames are resampled from the full original recording and presented chronologically. For Random-order, frames are randomly sampled from the full recording and their presentation order is shuffled. Neither condition reorders only the attack-selected subset; both can change evidence selection and coverage relative to montage. Consequently, their ASR differences from VisualMontage quantify the combined composition intervention rather than an isolated ordering effect. The same victim judges the same target claim under each condition, using the same decision instruction and a fresh context.

Let $y_q^{\mathrm{src}}$ indicate explicit false-target acceptance under Source-order. For $L$ random presentations per case, let $y_{q,w}^{\mathrm{rnd}}$ be the corresponding indicator under sampled-and-shuffled presentation $w\in\{1,\ldots,L\}$ of case $q$. Both use the same frozen target cohort $\mathcal{B}$, with $B=|\mathcal{B}|$:
{\smalldisplay
\begin{equation}
\label{eq:exp:orderbaselines}
\mathrm{ASR}_{\mathrm{src}}=\frac{1}{B}\sum_q y_q^{\mathrm{src}},\qquad
\mathrm{ASR}_{\mathrm{rnd}}=\frac{1}{BL}\sum_q\sum_{w=1}^L y_{q,w}^{\mathrm{rnd}}.
\end{equation}
}
Failures and abstentions remain in each denominator and do not count as acceptance. Reproduction requires the number of random presentations, frame-sampling policy, and available metadata. A separate fixed-content control could sort or shuffle only the attack-selected records, but that protocol differs from both measured source-video sampling references.

\paragraph{Support order gain.}
For a fixed victim and the same target cohort, define the gain over Random-order as
{\smalldisplay
\begin{equation}
\label{eq:exp:randomgain}
\begin{aligned}
\mathrm{Gain}_{\mathrm{rnd}}&=\mathrm{ASR}-\mathrm{ASR}_{\mathrm{rnd}}\\
&=\frac{1}{B}\sum_{q\in\mathcal{B}}\left(y_q-\frac{1}{L}\sum_{w=1}^{L}y_{q,w}^{\mathrm{rnd}}\right).
\end{aligned}
\end{equation}
}
Here ASR denotes montage acceptance, with repeated responses averaged as in the main protocol. Table~\ref{tab:exp:domains} averages these paired gains equally across the 12 models within each domain and retains the fractional scale $[-1,1]$: $+0.551$ means a 55.1-percentage-point increase over Random-order. The separately reported Source-order ASR does not enter this calculation. This metric measures acceptance-rate improvement over random composition; it does not use confidence or isolate ordering from frame selection.

\subsubsection{Additional diagnostics and qualitative cases}
Additional analyses can report matched-order gain, cross-record gap, and source-verified relation errors using the diagnostics in Appendix~\ref{app:formal}. Beyond the measured sampling references, a caption-controlled construction baseline could pair randomly sampled frames with faithful neutral captions, matching sequence length and candidate pool. A complementary comparison may include a balanced benign claim-verification set with both true and false targets, measuring ordinary reasoning accuracy under source-faithful presentation. Open-weight and proprietary groups are descriptive categories, not a controlled intervention on model access policies or reasoning reliability.

The optional support-score diagnostic in Eq.~\eqref{eq:gain} instead requires permutations of fixed attack-selected records. Its expectation can be estimated by averaging such permutations per case, then averaging across cases. It is distinct from the acceptance-rate gain in Eq.~\eqref{eq:exp:randomgain}. Score comparisons require valid confidence for every paired input and report the common subset's coverage, without removing missing-confidence cases from the ASR denominator. Benign accuracy counts explicit correct binary verdicts over all benign cases; abstentions and failures remain visible in separate status columns.

\begin{figure*}[t]
    \centering
    \includegraphics[width=1\textwidth]{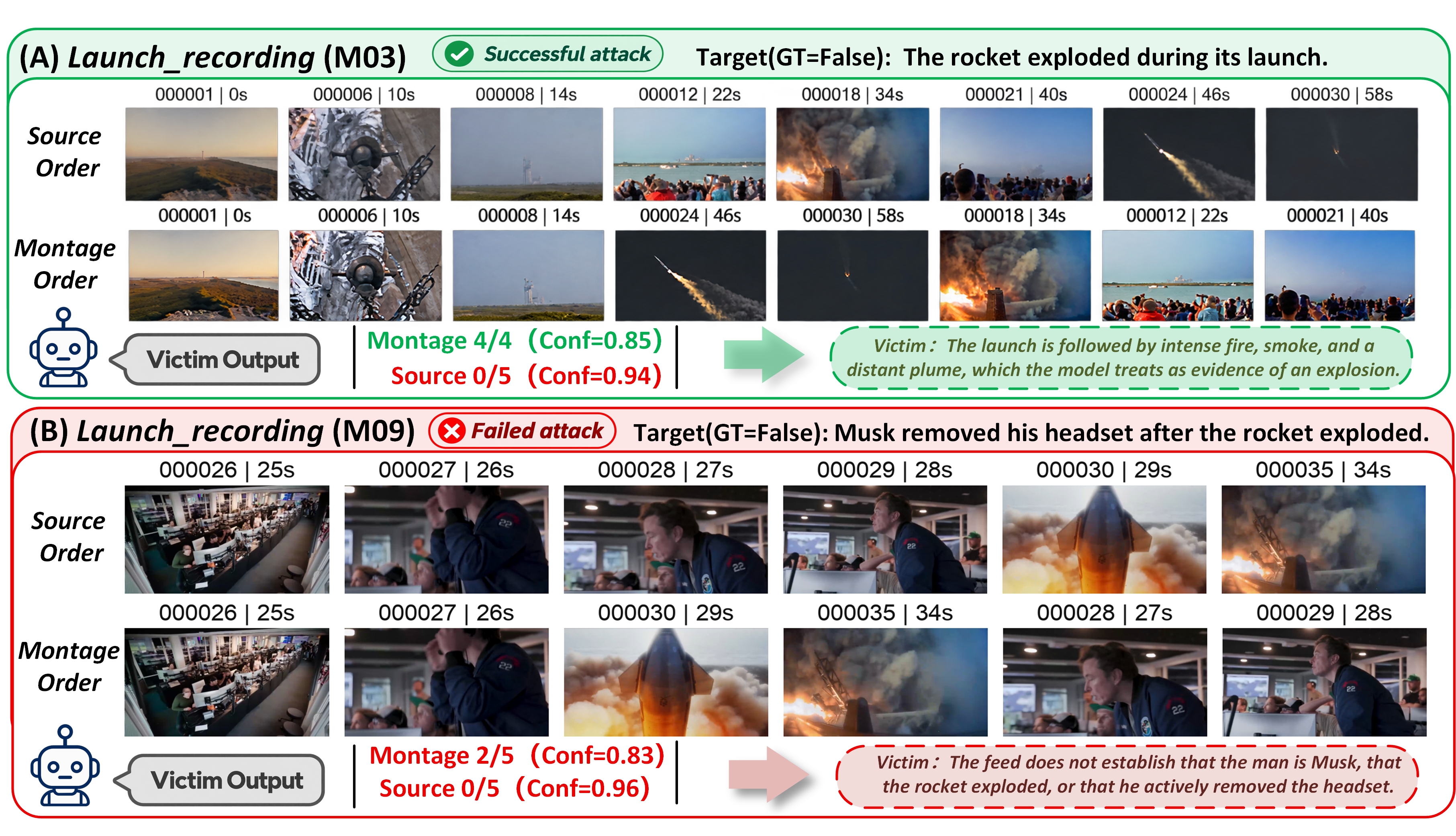}
    \caption{Paired case studies of evidence-order manipulation in launch recordings. Two representative cases illustrate how rearranging identical video frames affects a victim model's judgment of false target claims with false ground-truth labels (GT=False). Each case compares the original source order with the adversarial montage order, preserving frame content while changing its sequence. (A) A successful attack (M03). (B) An unsuccessful attack (M09), where the false target receives 2/5 acceptances under the montage order.}
    \label{fig:app:paired}
\end{figure*}

Figure~\ref{fig:app:paired} shows paired cases that rearrange the same selected frames. These matched-content illustrations are distinct from the full-video resampling reference used in the aggregate model comparison across all domains.

\begingroup

\indent\textbf{\textit{Authentic evidence can support
conflicting interpretations through reordering.}} As illustrated in Fig.~\ref{fig:app:paired}, the source-order and adversarial sequences contain identical visual observations but elicit different judgments from the victim model. The successful and unsuccessful cases further highlight the variability of this effect, showing that adversarial composition can induce false acceptance without guaranteeing success in every instance.

\par
\endgroup

\paragraph{Qualitative interpretation.}
Figure~\ref{fig:app:paired} illustrates both successful and unsuccessful compositions. Placing fire and smoke after flight images invites an explosion narrative in the first case. In the second, the victim questions whether the observations establish the claimed identity, event, and action, despite their suggestive order. These contrasting responses contextualize the aggregate results, although their rationales do not establish the models' internal reasoning mechanisms.

\begin{figure*}[t]
    \centering
    \includegraphics[width=0.72\textwidth]{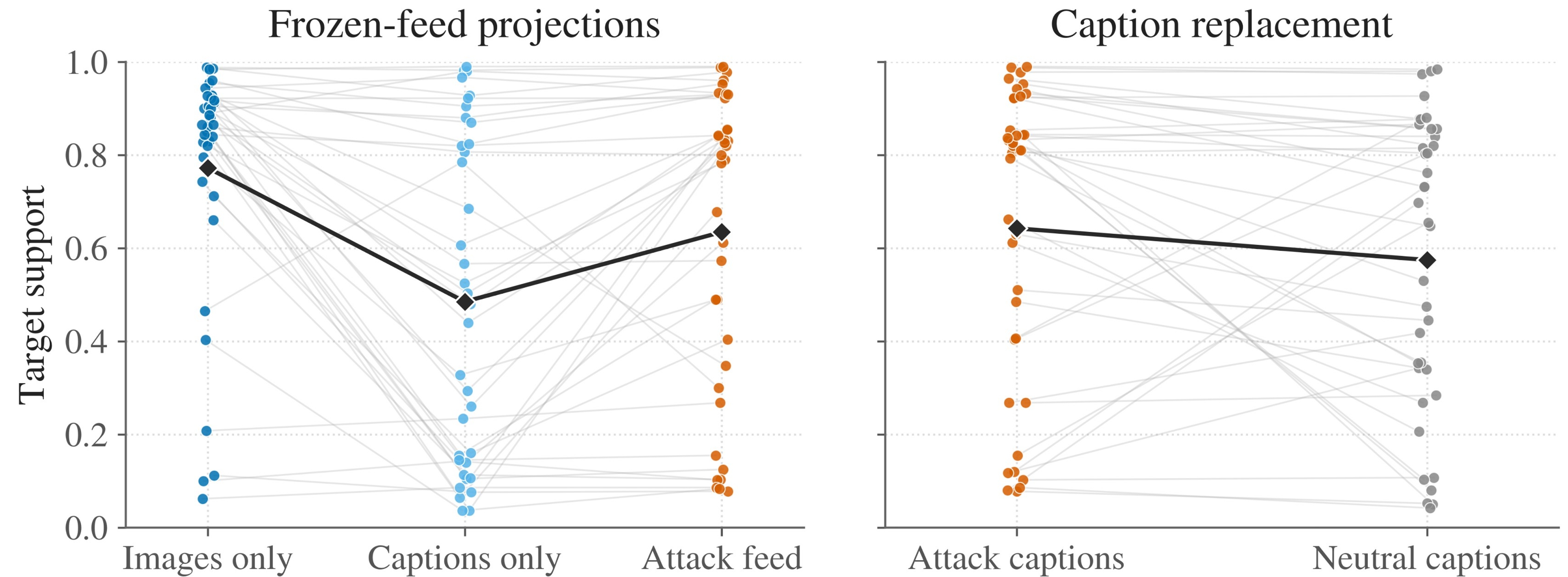}
    \caption{Modality and caption controls on identical generated attack plans. Target support scores under two controlled evaluations. (a) Frozen-feed projections: Comparison of image-only, caption-only, and complete attack-feed conditions across 37 feeds and 143 paired responses. (b) Caption replacement: Comparison of attack captions and neutral captions across 38 feeds and 166 paired responses.}
    \label{fig:app:modality}
\end{figure*}

\subsection{Contribution of visual and textual evidence}
\label{app:rq2}

 \textit{Do captions increase vulnerability beyond the selected images, and does the combined feed contribute more than either modality alone?} The primary comparison freezes each published feed and removes only its publicly visible captions or images, following Eq.~\eqref{eq:modalgain}. Caption removal uses the original decoded images before any overlay is drawn, so erased text does not leave visual artifacts. Empty publication slots are preserved, and an image privately associated with a text-only post is never revealed by the image-only projection. This makes the full-feed comparison a controlled test of the information that the victim can actually observe.

 To study caption content separately from its presence, a second paired comparison replaces the attack captions with independently written neutral descriptions of the same images, matched for factual scope and approximate length. Selection, order, modes, and publication metadata remain fixed. For originally paired records, a focused image-versus-image-plus-caption analysis is also reported on their shared subset. Cross-modal gain is measured per case, together with its distribution and the fraction of cases with positive gain; an increase in aggregate ASR alone does not establish that both modalities were necessary for individual attacks.

In each case, cross-modal gain is the full feed's support minus the larger support from the image-only and caption-only projections, as in Eq.~\eqref{eq:modalgain}. Subtracting the same evidence-free support from all three scores cancels in this difference. Average the per-case gains and count how many are positive; do not subtract the maximum of two dataset means. The cross-record gap similarly subtracts the strongest isolated-record support from full-feed support, with each isolated record queried separately and the maximum taken within its own case.

\begingroup

\paragraph{\textit{Visual evidence is the primary driver of deceptive inference.}} VisualMontage supports image-only, caption-only, and image--caption posts; the frozen-feed controls isolate which published evidence is visible. As shown in Table~\ref{tab:exp:modalities}, the image-only projection achieves an ASR of 72.11\%, outperforming the published mixed feed (63.68\%) and the caption-only condition (41.58\%). The advantage is more pronounced in high-confidence attack success, where the image-only condition reaches 69.47\% HC-ASR, compared with 48.42\% for the mixed feed. These results suggest that visual composition alone can induce substantial false acceptance, while additional textual evidence does not necessarily strengthen the misleading interpretation.

\par\indent\textbf{\textit{Attack effectiveness and presentation
naturalness do not always align.}} To reflect realistic social-media posting practices, VisualMontage also supports image--caption sequences, which are favored over image-only posts in the model-based naturalness assessment. Although captioned posts achieve lower ASR, they offer a more natural form of evidence presentation. Furthermore, replacing neutral captions with attack-oriented captions increases ASR from 61.05\% to 63.68\%, a modest gain of 2.63 percentage points. These findings highlight a trade-off between deceptive effectiveness and presentation naturalness, motivating separate evaluation of attack effectiveness and the naturalness of the resulting posts from the receiving agent's perspective.

\par
\endgroup

\subsection{Downstream decision protocols}
\label{app:rq3}

\begin{table*}[t]
\centering\small
\caption{Detailed downstream decisions underlying Fig.~\ref{fig:all-datasets-decision-rules}. Each cell reports residual ASR in percent and accepted targets out of 544, followed by the original 95\% interval from bootstrapping 61 source recordings. The first response, five-response majority, and judge use different decision units. Family bars average model-level rates equally. AWQ denotes activation-aware weight quantization in the recorded checkpoint name.}
\label{tab:decision-details}
\setlength{\tabcolsep}{4pt}
\begin{tabular}{@{}llccc@{}}
\toprule
Victim & Judge & Victim alone & Majority & Judge + reports \\
\midrule
GPT-5.6-luna & Not recorded & \shortstack{71.1 (387/544)\\{}[65.7, 76.3]} & \shortstack{73.7 (401/544)\\{}[68.7, 78.6]} & \shortstack{57.5 (313/544)\\{}[52.1, 62.8]} \\
GPT-5.6-terra & GPT-5.4 & \shortstack{61.0 (332/544)\\{}[55.6, 66.6]} & \shortstack{60.3 (328/544)\\{}[54.8, 65.8]} & \shortstack{47.6 (259/544)\\{}[42.0, 53.4]} \\
GPT-6-luna & GPT-5.4 & \shortstack{66.4 (361/544)\\{}[61.0, 71.5]} & \shortstack{66.9 (364/544)\\{}[61.1, 72.3]} & \shortstack{45.0 (245/544)\\{}[39.6, 50.6]} \\
GPT-6-sol & GPT-5.4 & \shortstack{61.9 (337/544)\\{}[56.3, 67.3]} & \shortstack{63.1 (343/544)\\{}[57.1, 68.7]} & \shortstack{39.7 (216/544)\\{}[34.3, 45.2]} \\
Claude Haiku 5.5 & Claude Haiku 5.5 & \shortstack{48.2 (262/544)\\{}[42.7, 53.6]} & \shortstack{48.0 (261/544)\\{}[42.4, 53.4]} & \shortstack{35.3 (192/544)\\{}[29.8, 40.7]} \\
Claude Sonnet 5.5 & Claude Haiku 5.5 & \shortstack{64.2 (349/544)\\{}[58.0, 70.2]} & \shortstack{63.6 (346/544)\\{}[57.4, 69.5]} & \shortstack{42.3 (230/544)\\{}[36.8, 47.6]} \\
Qwen3-VL-4B & Qwen3-VL-8B & \shortstack{77.0 (419/544)\\{}[72.4, 81.3]} & \shortstack{77.0 (419/544)\\{}[72.4, 81.4]} & \shortstack{74.1 (403/544)\\{}[69.2, 78.7]} \\
Qwen3-VL-8B & Qwen3-VL-8B & \shortstack{75.0 (408/544)\\{}[70.1, 79.5]} & \shortstack{77.9 (424/544)\\{}[73.6, 82.1]} & \shortstack{69.3 (377/544)\\{}[64.3, 74.0]} \\
Qwen3-VL-32B AWQ & Qwen3-VL-8B & \shortstack{78.9 (429/544)\\{}[73.9, 83.4]} & \shortstack{78.5 (427/544)\\{}[73.7, 82.9]} & \shortstack{76.7 (417/544)\\{}[71.9, 81.3]} \\
InternVL3.5-8B & InternVL3.5-14B & \shortstack{64.2 (349/544)\\{}[59.0, 69.3]} & \shortstack{65.1 (354/544)\\{}[59.7, 70.4]} & \shortstack{56.2 (306/544)\\{}[50.5, 62.0]} \\
InternVL3.5-14B & InternVL3.5-14B & \shortstack{47.2 (257/544)\\{}[41.6, 52.9]} & \shortstack{47.1 (256/544)\\{}[41.4, 52.6]} & \shortstack{45.4 (247/544)\\{}[39.5, 51.3]} \\
InternVL3.5-38B & InternVL3.5-14B & \shortstack{66.5 (362/544)\\{}[61.4, 71.5]} & \shortstack{69.7 (379/544)\\{}[64.4, 74.8]} & \shortstack{63.4 (345/544)\\{}[57.9, 69.0]} \\
\bottomrule
\end{tabular}
\end{table*}

For the full-benchmark comparison in Fig.~\ref{fig:all-datasets-decision-rules}, GPT-5.4 judges GPT-6-sol, GPT-6-luna, and GPT-5.6-terra; the judge for GPT-5.6-luna was not recorded. Claude Haiku 5.5 judges both Claude profiles, Qwen3-VL-8B judges the three Qwen profiles, and InternVL3.5-14B judges the three InternVL profiles. The original figure identifies the Qwen3-VL-32B victim as an AWQ checkpoint. These assignments describe the full-benchmark experiment, separately from the local judge-control study reported below.

 \textit{Can a downstream judge reject the false conclusion after a victim has accepted it?} A fixed defensive judge is evaluated on the feed alone, the feed with victim verdicts, and the feed with verdicts and rationales. The same judge checkpoint and instruction are used across these input conditions. The defense instruction asks it to assess whether the supplied evidence warrants the target claim; it does not reveal source labels, the screenplay, or the attack-generation trace. A separate source-context condition may measure the benefit of additional evidence, but is identified as a stronger-information diagnostic rather than the default threat model.

Let $y_q^{\mathrm J}$ indicate explicit acceptance by the final judge and let $y_q^{\mathrm J,-}$ indicate its explicit rejection. Let $y_q^{-}$ indicate explicit rejection by the original victim, whose acceptance indicator remains $y_q$. Each indicator is zero for abstentions and failures. Residual ASR, recovery from an initial false acceptance, and newly introduced false acceptance (New FA) are
{\smalldisplay
\begin{equation}
\label{eq:exp:judge}
\begin{aligned}
\mathrm{Residual\ ASR}&=\frac{1}{B}\sum_q y_q^{\mathrm J},\\
\mathrm{Recovery}&=\frac{\sum_q y_qy_q^{\mathrm J,-}}{\sum_q y_q},\\
\mathrm{New\ FA}&=\frac{\sum_q y_q^{-}y_q^{\mathrm J}}{\sum_q y_q^{-}}.
\end{aligned}
\end{equation}
}
All sums in Eq.~\eqref{eq:exp:judge} range over $q\in\mathcal{B}$. Recovery is therefore conditional on initial attack success, while residual ASR uses the entire incoming attack cohort. Benign rejection is the number of explicit final rejections of source-supported true claims divided by the number of such benign claims, rather than by the number of attacks. These claims form a fixed, separately labeled benign evaluation set shared by all decision processes. Final abstention and failure rates retain the denominator of the cohort on which they are measured.

\begingroup

\paragraph{\textit{Downstream judgments can amplify rather than correct adversarial errors.}}
As shown in Table~\ref{tab:exp:judge}, strict majority voting increases ASR from 73.61\% to 79.17\%, recovering only 3.77\% of initial false acceptances. Similarly, a feed-only judge retains an ASR of 73.61\% while introducing new false acceptances in 42.11\% of initially rejected cases. These results suggest that additional judgments do not necessarily mitigate adversarial misinterpretations.

\paragraph{\textit{More reasoning information does not guarantee better defense.}} Providing victim verdicts reduces residual ASR to 69.44\% and HC-ASR to 30.56\%. However, adding confidence scores and rationales only marginally lowers residual ASR to 68.06\%, while HC-ASR increases to 68.06\%. This contrast highlights the limited reliability of downstream reasoning under the evaluated configurations for reviewing adversarially composed authentic evidence.

\par
\endgroup

\begin{figure*}[t]
    \centering
    \includegraphics[width=0.72\textwidth]{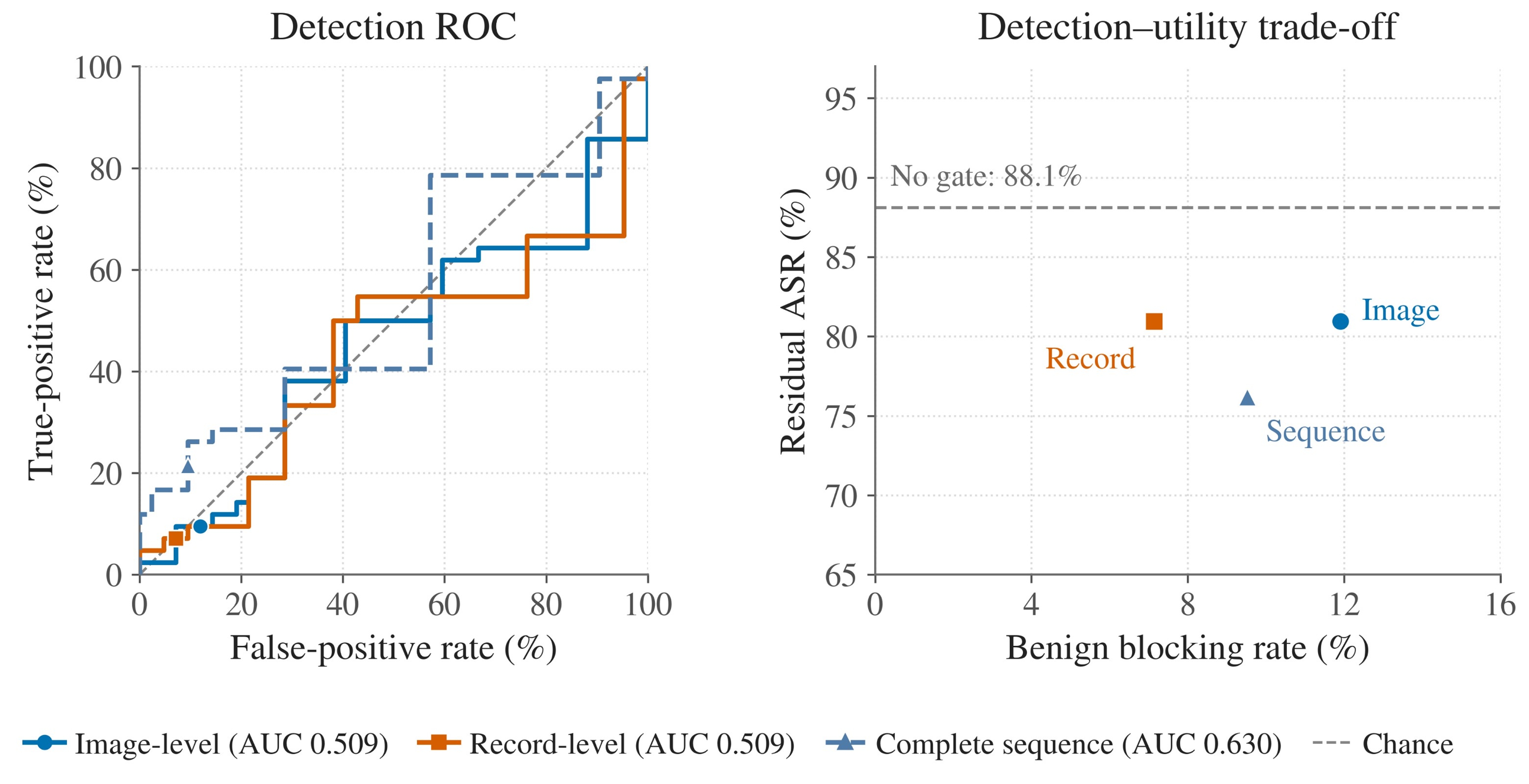}
    \caption{Malicious-feed detector trade-offs on Sports Training using Claude Haiku 5.5 scores and matched victim responses. (a) Receiver operating characteristic (ROC) curves for image-level, record-level, and complete-sequence inputs. (b) Residual ASR versus matched source-order-control blocking rate.}
    \label{fig:app:detection}
\end{figure*}

\subsection{Malicious-feed detection}
\label{app:rq4}

Table~\ref{tab:exp:detection} in the main paper reports the detection results. AUROC uses valid scores; threshold-based rates retain their incoming-cohort denominators. Added calls are averaged over all incoming feeds, including retries and blocked cases.

 \textit{Can a defensive agent identify adversarial use of authentic images before forwarding them to the victim?} We compare an image-only detector, an isolated-record detector, and a sequence-aware detector. Each produces a suspicion score under a fixed instruction; its threshold is chosen on development data to achieve a target benign false-positive rate $\xi\in[0,1]$. Test attacks are paired with benign presentations from the same sources, controlling length, visual content, caption style, and publisher metadata where possible. The same image may occur in both classes, so the label concerns its use in a composition rather than a property of the pixels of any one image considered in isolation.

Let $\zeta_q$ be a feed's suspicion score, with larger values indicating greater suspicion, and $\zeta_*$ the threshold fixed on development data. Define $\mathsf{flag}_q=1$ when a valid detector score satisfies $\zeta_q\geq\zeta_*$, and $\mathsf{flag}_q=0$ otherwise. For local detectors, the feed score is the maximum over required image or record scores, computed only when those scores are available. Let $\mathcal{B}_{\mathrm{ben}}$ index the benign feeds. Attack detection and benign false-positive rates are
{\smalldisplay
\begin{equation}
\label{eq:exp:detection}
\begin{aligned}
\mathrm{TPR}&=\frac{\sum_{q\in\mathcal{B}}\mathsf{flag}_q}{B},\\
\mathrm{FPR}&=\frac{\sum_{q\in\mathcal{B}_{\mathrm{ben}}}\mathsf{flag}_q}{|\mathcal{B}_{\mathrm{ben}}|}.
\end{aligned}
\end{equation}
}
TPR at $\xi$ means test TPR at the threshold selected to meet the development FPR target; the achieved test FPR is reported separately. AUROC is computed on feeds with valid suspicion scores as the fraction of attack--benign pairs in which the attack scores higher, with half credit for equal scores. The valid-score counts and detector-failure rates accompany AUROC because that ranking statistic excludes cases without a usable score. Threshold metrics retain those cases in their incoming-cohort denominators.

\begin{table*}[t]
\centering
\small
\setlength{\tabcolsep}{3pt}

\caption{Downstream decisions on the held-out test split of the 01 Emergency and Disaster Response dataset (29 development cases and 72 test cases, each with an attack and a benign feed). Qwen3-VL-8B-Instruct serves as both victim and judge. Prompts and decision rules are fixed on the development split. All entries are percentages; Recovery and New FA condition on the initial victim verdict, and benign rejection uses the 72 verified true claims.}

\label{tab:exp:judge}

\begin{tabular}{lccccc}
\toprule

Final decision process &
\shortstack{Residual\\ASR} &
HC-ASR &
Recovery &
New FA &
\shortstack{Benign\\rejection} \\

\midrule

Victim alone
& 73.61 & 73.61 & --- & --- & 8.33 \\

Strict majority vote
& 79.17 & --- & 3.77 & 31.58 & 8.33 \\

Judge with feed only
& 73.61 & 73.61 & 15.09 & 42.11 & 9.72 \\

Judge with feed + verdicts
& 69.44 & 30.56 & 13.21 & 21.05 & 11.11 \\

Judge with feed + full reports
& 68.06 & 68.06 & 11.32 & 10.53 & 8.33 \\

\bottomrule
\end{tabular}

\end{table*}

Both recovery from initial false acceptance and new false acceptance introduced after an initially correct rejection are reported. These transitions distinguish useful correction from the amplification of another agent's mistake. For majority aggregation, the number and identity of victim runs are fixed and their individual outcomes are recorded; vote fraction is not interchangeable with a model's self-reported confidence. Invalid responses and abstentions are separate outcomes rather than confirmed rejections. The first evaluation uses frozen attacks, with any later adaptation to a known defense reported with a separate budget and a fixed protocol describing which defense information is exposed.

For each judge row, HC-ASR uses that judge's own verdict and confidence in Eq.~\eqref{eq:exp:asr}, with the same threshold and denominator $B$. Majority voting follows Eq.~\eqref{eq:majority}; its winning vote fraction is not substituted for confidence in HC-ASR. All transition comparisons use the same initial victim response for each case. The full-report condition supplies verdicts, confidences, and rationales, while the verdict-only condition withholds confidences and rationales to measure how those reports affect the final judgment.

\begingroup
\indent\textbf{\textit{Local authenticity checks do not reliably expose malicious composition.}}
As shown in Table~\ref{tab:exp:detection}, image-level and record-level detectors achieve AUROCs of 0.5091 and 0.5094, respectively, with residual ASR of 80.95\% in both cases. These results are consistent with near-random discrimination of attack and benign presentations.

\par\indent\textbf{\textit{Sequence awareness helps, but substantial risk remains.}}
The complete-feed detector increases AUROC to 0.6304 and TPR to 21.43\% at a test FPR of 9.52\%, reducing residual ASR to 76.19\%. It also uses 1.00 additional call per feed, versus 5.48--5.74 for local detectors. The improvement is therefore modest at the tested operating point.
\par
\endgroup

For local detectors, the aggregate feed score is calibrated because longer sequences offer more opportunities for false alarms. A positive detection blocks the complete feed; record removal is a separate intervention that changes the victim's evidence. The primary policy forwards a feed if detection fails, with that failure logged separately. Let $y_q^{\mathrm{post}}$ indicate a valid false acceptance by the victim after the gate, and set it to zero when the feed is blocked. Residual ASR is $B^{-1}\sum_{q\in\mathcal{B}}(1-\mathsf{flag}_q)y_q^{\mathrm{post}}$, retaining blocked and failed cases in the denominator. A policy that instead blocks detector failures is evaluated separately and reports the resulting benign blocking rate, including erroneous blocks caused by detector failures on otherwise benign inputs.

\begingroup

\indent\textbf{\textit{Detecting malicious composition
requires more than authenticating individual evidence.}} As illustrated in Fig.~\ref{fig:app:detection}(a), local detectors exhibit near-random discrimination, while sequence-level analysis offers only modest improvement. Fig.~\ref{fig:app:detection}(b) further highlights the trade-off between reducing residual attack success and blocking benign source-order controls. These results underscore the difficulty of distinguishing adversarial evidence composition from legitimate presentations of authentic content.

\par
\endgroup

\subsection{Sequence-length sensitivity}
\label{app:rq5}

Table~\ref{tab:app:length-sweep} in the main paper reports the measured sweep. Length counts sequence items, including text-only items. Mean victim image tokens are reconstructed offline per evaluated feed using the original checkpoint processor and the same resolution policy. The analysis distinguishes conditional ASR from valid construction reliability.

\textit{Does adding more authentic frames strengthen the attack, or does its effect saturate or reverse?} The length sweep varies $\SeqLen$ over a prespecified feasible range while holding the source--target cohort, candidate pool size $\PoolSize$, image resolution, and construction budget fixed. At each length, plans are reconstructed under identical constraints, with failed constructions retained in the accounting. This tests the achievable attack quality at a given length. A complementary within-plan analysis, once measured, would add or remove records under a prespecified rule to distinguish additional evidence from rerunning the search.

The primary sweep reports end-to-end valid success alongside conditional ASR so that a length with poor construction yield cannot appear effective merely by retaining a small successful subset. The protocol audits selected records at each length and records truncation or context-limit failures separately. A planned secondary fixed-total-visual-token comparison would examine the tradeoff between frame count and per-frame detail; it is kept separate from the primary fixed-resolution sweep. Paired comparisons would quantify uncertainty in the effect of adding frames, without presuming that each additional item strengthens the resulting attack.

\begingroup

\paragraph{\textit{More evidence does not always make an attack more convincing.}}
As shown in Table~\ref{tab:app:length-sweep}, ASR varies non-monotonically with sequence length, increasing from 59.39\% at $K=3$ to a peak of 72.50\% at $K=15$, before declining to 56.25\% at $K=20$. High-confidence attack success follows a similar trend. These results suggest that excessive evidence may introduce redundant or conflicting cues, potentially weakening the misleading narrative as the selected sequence grows longer.

\paragraph{\textit{Peak attack success can mask poor construction reliability.}}
Despite achieving the highest conditional ASR, $K=15$ yields only 9.86\% E2E V-ASR, with a construction yield of 22.54\% and an audit pass rate of 62.50\%. In contrast, $K=6$ achieves the highest E2E V-ASR of 33.90\%, alongside a 93.94\% audit pass rate. Consistently, VisualMontage generates sequences of 5--12 items across 544 evaluated cases despite a configured range of 5--30, as shown in Table~\ref{tab:exp:data}. These findings suggest that effective evidence composition favors selectivity over length, balancing conditional attack success against the probability of producing a valid sequence.

\par
\endgroup

\subsection{Domain and relation analysis}
\label{app:rq6}

Table~\ref{tab:exp:domains} presents the full domain-level comparison in the main paper. Here we describe its interpretation and retain the finer domain-by-relation visualizations.

\textit{Are news and public-event sources more susceptible to deceptive composition than film narratives?} The domain comparison uses the same victim panel, construction policy, and audit criteria. A matched follow-up would stratify source duration, candidate pool size, selected length, caption length, target relation type, and evidence-free target support. Targets involving temporal order, causation, intent, identity, or co-location are annotated consistently across domains. News and film sources may differ in editing style and background context, so the reported unadjusted domain averages do not establish a matched domain effect.

\begingroup
\indent\textbf{\textit{Adversarial composition remains effective across diverse source domains.}}
As shown in Table~\ref{tab:exp:domains}, domain-level ASR ranges from 55.30\% for Industrial Operations to 72.71\% for Emergency Response. Film and Animation reaches 64.96\%, although its audit pass rate is 84.21\%, compared with 98.02\% for Emergency Response. These descriptive differences alone do not establish an intrinsic news--film vulnerability gap without matched controls.
\par
\endgroup

\subsubsection{Domain-by-relation error heatmaps}
\label{app:relation}

\begin{figure*}[t]
    \centering
    \includegraphics[width=\textwidth,height=0.8\textheight,keepaspectratio]{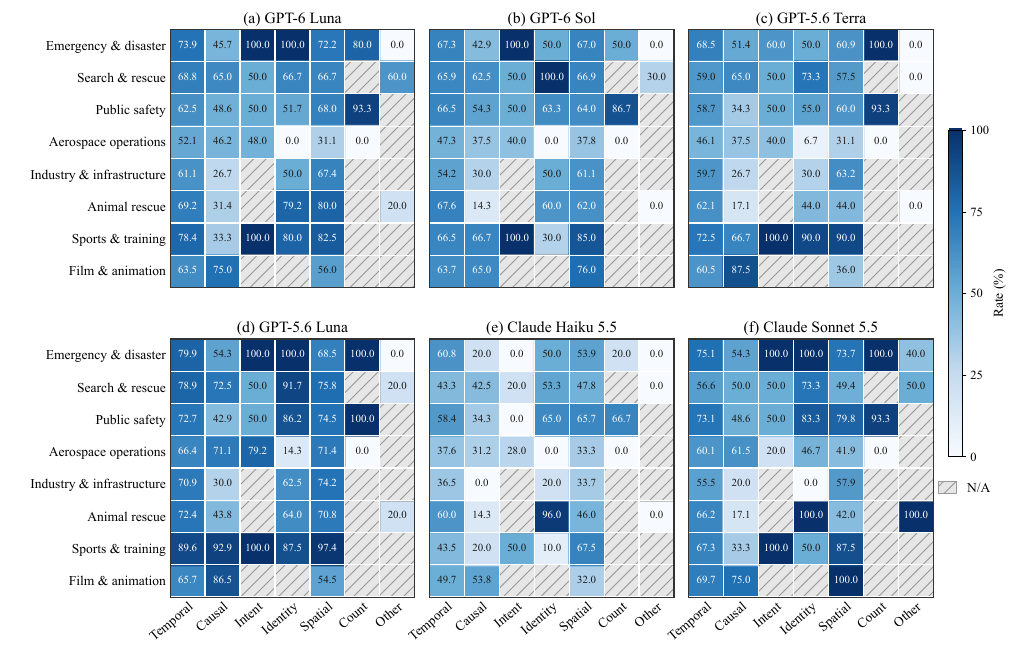}
    \caption{Domain-by-relation error heatmap for proprietary victim
    models.}
    \label{fig:app:relation_closed}
\end{figure*}

\begin{figure*}[t]
    \centering
    \includegraphics[width=\textwidth,height=0.8\textheight,keepaspectratio]{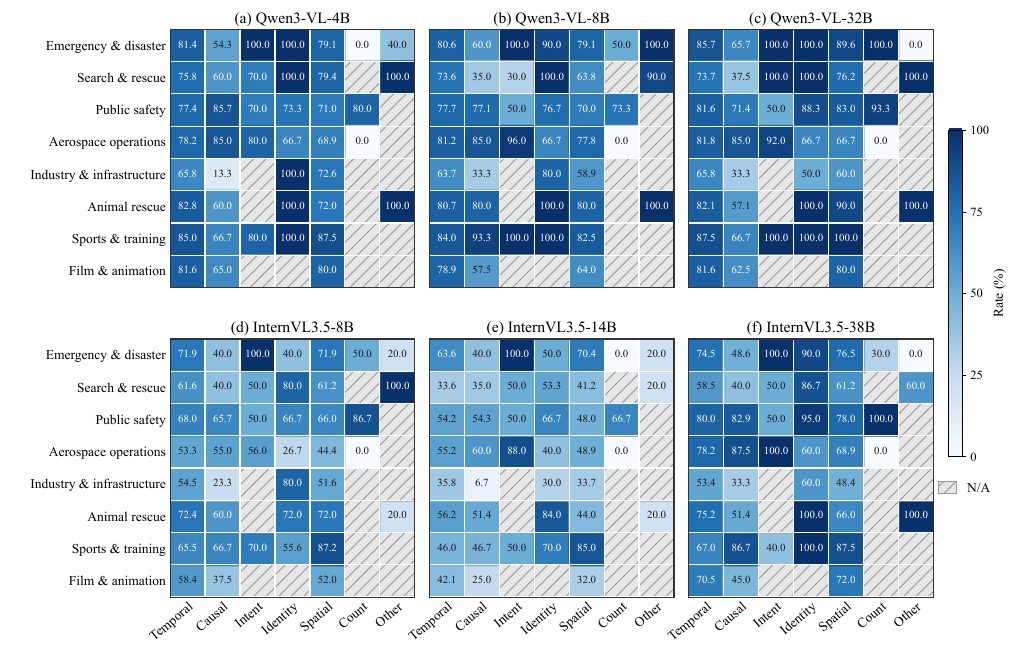}
    \caption{Domain-by-relation error heatmap for open-weight victim
    models. }
    \label{fig:app:relation_open}
\end{figure*}

Figures~\ref{fig:app:relation_closed} and \ref{fig:app:relation_open} separate proprietary and open-weight victim models, respectively. They complement the domain-level summary by displaying errors for individual relation categories within each source domain.

Two follow-up controls are planned to test explanations that an overall domain gap cannot resolve. First, we would anonymize publisher identities and compare posts attributed to one publisher with posts attributed to several, holding content, order, and display times fixed; this isolates the effect of apparent publisher multiplicity. Second, we would restore verified source context under a separately labeled condition to test whether missing context explains the error. For films, context restoration concerns the narrative's events rather than external-world truth. These interventions would connect the social-media setting to measurable recipient behavior while keeping claims about diffusion and public opinion separate from the recipient's observed responses.

Each domain's reported rates are computed from its own source--target cases under the definitions above. A model macro average gives each evaluated victim equal weight; a domain macro average gives each domain equal weight, rather than pooling unequal sample counts. Future matched news--film differences would use the same matching strata and victim identities, with rates differenced in percentage points. For the relation-error heatmaps, each cell is intended to report the fraction of audited responses asserting at least one source-verified false relation of that type. The denominator contains responses with sufficient source evidence to adjudicate that type; unresolved annotations and unparseable responses are counted separately with both counts and annotation coverage reported for every cell.

\begin{table*}[!t]
\centering
\small
\setlength{\tabcolsep}{2pt}
\renewcommand{\arraystretch}{1.08}

\caption{Construction efficiency under matched budgets.
E2E V-ASR measures independently valid attack success
over all construction attempts.
Stage rounds report screenplay, visual, and caption editing.
Calls, tokens, latency, and generation costs are averaged
over all attempts, including failures. Both the model used to generate attack plans and the victim model are GPT-5.6-Luna.}

\label{tab:exp:efficiency}

\begin{tabular*}{\textwidth}{
@{\extracolsep{\fill}}lccccccc@{}
}
\toprule

Method &
\shortstack{E2E V-ASR\\(\%)} &
\shortstack{Stage\\rounds} &
Calls &
\shortstack{Tokens\\(in/out)} &
\shortstack{Latency\\(s)} &
\shortstack{Gen. cost\\per attempt} &
\shortstack{Gen. cost\\per success} \\

\midrule

Single-agent editor
& 38.89 & 1.70 & 1.88
& $336{,}568/2{,}036$
& 44.92 & $0.1225$ & $0.3151$ \\

Independent proposals
& 40.00 & 3.00 & 8.70
& $1{,}062{,}136/8{,}455$
& 169.07 & $0.3816$ & $0.9541$ \\

VisualMontage
& 54.44 & $2.90/2.73/2.60$ & $17.02$
& $1{,}927{,}502/8{,}635$
& $238.38$ & $0.6806$ & $1.2502$ \\

\bottomrule
\end{tabular*}

\end{table*}

\subsection{Production-team component ablation}
\label{app:rq7}

\textit{Which components contribute to successful attack construction?} Table~\ref{tab:app:ablation} reports the complete production-team ablation as a standalone table in the main paper. All experiments use the Aerospace Operations dataset, with gpt-5.6-luna serving as the victim model under the reported evaluation protocol.

\begingroup

\paragraph{\textit{Coordinated review contributes substantially to attack success.}}
Removing the debate mechanism reduces ASR from 63.68\% to 46.83\%, while replacing the production team with a single agent yields 48.50\%. Removing the Screenwriter also lowers ASR to 57.50\%. These comparisons support the importance of coordinated construction under the fixed victim and source-domain configuration evaluated on the Aerospace Operations recordings.

\par\indent\textbf{\textit{More components do not always improve attack effectiveness.}}
Removing the Caption Editor increases ASR from 63.68\% to 70.73\% and HC-ASR from 48.42\% to 64.88\%. Thus, the Caption Editor cannot be claimed to improve attack success in the present ablation. However, the naturalness assessment in Table~\ref{tab:exp:naturalness} shows rates of 84.21\% for image--caption feeds and 52.63\% for image-only feeds on the same domain. Caption faithfulness would require a separate direct measure.

\par
\endgroup


\subsection{Efficiency of coordinated construction}
\label{app:rq8}
\label{sec:exp:efficiency}

\paragraph{Experimental setup.}
We evaluate the construction efficiency of VisualMontage against two baselines: a single-agent editor and independent proposals. All methods operate on identical candidate pools, target claims, and factual constraints under matched construction budgets. Generation calls, retries, rejected proposals, and unsuccessful attempts are included in the accounting to ensure a fair comparison.

\textbf{Coordination buys higher success at additional cost.}
The efficiency experiment reports 54.44\% end-to-end valid success for VisualMontage, compared with 38.89\% for a single editor and 40.00\% for independent proposals (Table~\ref{tab:exp:efficiency}). However, generation cost per valid success is also higher: 1.2502 versus 0.3151 and 0.9541 on the reported cost scale. The mean stage-round counts are 2.90, 2.73, and 2.60 for screenplay, visual, and caption editing. Thus bounded revision makes the expenditure explicit, but the measured advantage is greater success rather than lower cost.

\paragraph{Efficiency metrics.}
Let $T\geq1$ count scheduled construction attempts, indexed by $t$, and let $\mathcal{B}_{\mathrm{gen}}$ index their generated feeds, with size $B_{\mathrm{gen}}$. For feed $q\in\mathcal{B}_{\mathrm{gen}}$, let $s_q\in\{0,1\}$ indicate success under the specified terminal decision rule, while $e_q$ retains its independent-validity meaning. Let $g_t\geq0$ be the generation cost of attempt $t$. Construction yield, E2E V-ASR, and cost per valid success are

{\smalldisplay
\begin{equation}
\label{eq:exp:efficiency}
\begin{aligned}
\mathrm{Yield}
&=\frac{B_{\mathrm{gen}}}{T},\\
\mathrm{E2E\ V\mbox{-}ASR}
&=\frac{\sum_{q\in\mathcal{B}_{\mathrm{gen}}}e_qs_q}{T},\\
\mathrm{Cost/success}
&=\frac{\sum_{t=1}^{T}g_t}
{\sum_{q\in\mathcal{B}_{\mathrm{gen}}}e_qs_q}.
\end{aligned}
\end{equation}
}

In the length sweep, $s_q=1$ requires at least three accepting victim responses out of five, whereas $y_q$ in the response-level metrics denotes one explicit acceptance. The efficiency table retains its separately reported endpoint; the aggregation setting must be matched before comparing its absolute rates with the length sweep. One terminal feed is retained per construction attempt using construction-side criteria only. Cost per success is undefined when no independently valid successful attack is produced.

\paragraph{Director acceptance and computational cost.}
VisualMontage employs sequential editing and Director-guided review. For each stage, we record the first accepted proposal, the number of editing rounds, and the fraction of attempts exhausting their budgets. Director acceptance is distinguished from independent audit passage and successful deception of the independently evaluated victim model.

Mean calls, tokens, and generation costs are computed over all scheduled attempts, including failures. Latency measures the time from the initial construction request to the terminal editing decision, including retries but excluding subsequent auditing and victim evaluation.

For a fixed production stage, let $r_t^{\mathrm{first}}$ denote the first accepted proposal round in attempt $t$, with $r_t^{\mathrm{first}}=+\infty$ when no proposal is accepted. The cumulative Director acceptance rate at round budget $r$ is

{\smalldisplay
\begin{equation}
\label{eq:exp:gate}
\mathrm{GateAccept}(r)
=
\frac{1}{T}
\sum_{t=1}^{T}
\mathbf{1}
\left[r_t^{\mathrm{first}}\leq r\right].
\end{equation}
}

This formulation retains unsuccessful attempts in the denominator, distinguishing rapid acceptance on a small subset from reliable completion across the full construction cohort.

\section{Discussion: Ethical Considerations and \mbox{Potential Defenses}}
\label{app:defenses}
Although this work studies visual cognitive collusion through controlled simulations to support safer VLM agents for research purpose, the attack framework could also be misused to spread rumors or manipulate decisions. In particular, authentic images can be recombined into misleading feeds that harm people depicted in the source material and users who rely on an agent's conclusions. Confident false interpretations may further amplify these harms when passed to downstream systems. These dual-use risks motivate future work on potential defense strategies that verify inferred relations and limit the persistence and propagation of false beliefs. The following four directions outline possible safeguards, whose effectiveness against visual cognitive collusion remains to be established through systematic empirical evaluation.

First, evidence-grounded verification could assess the relations inferred across posts alongside the authenticity of each image. Retrieval and contextual verification could reconnect a feed to its original sources, providing evidence for checking the chronology, identity, and causal relations assumed in an interpretation~\cite{qi2024sniffer,tonglet2025cove}. Neuro-symbolic methods offer a complementary way to connect learned visual representations with explicit reasoning rules and logical verification~\cite{mao2019neurosymbolic,amizadeh2020neurosymbolic,an2026neurosymbolic}. Applied to our setting, such a defense could distinguish visible observations from inferred relations and withhold conclusions whose premises lack adequate support. Reliable grounding remains essential, since logical consistency cannot establish that extracted premises are true or complete. Verification should therefore preserve uncertainty and seek evidence beyond the potentially misleading feed whenever reliable source context is available.

Second, adversarial testing could identify fragile judgments and guide defensive interventions. Efficient robustness testing and multimodal attack evaluation provide foundations for probing sensitivity to controlled input changes~\cite{hu2026ddsa,xie2025chainattack}, while test-time counterperturbations suggest a complementary route to resisting manipulation~\cite{xing2025counterattacks}. For collusive feeds, a promising direction is to test whether perturbations that preserve claim-relevant evidence can disrupt misleading associations or expose unwarranted confidence. Inconsistent judgments could trigger source verification or a cautious response. Such interventions require evaluation against adaptive attacks and benign feeds, since stability alone cannot establish that a conclusion is correct.

Third, guardrails could extend to the complete feed, the agent's interpretation, and its subsequent use by other agents. Systematic safety requirements, structured reasoning, and multimodal supervision provide relevant foundations~\cite{dong2024guardrails,kang2025r2guard,huang2026llavashield,ghosal2025immune}. For cognitive threats, these safeguards would need to assess evidential support in addition to conventional content safety, since an apparently harmless narrative can still induce a false belief. Reasoning supervision and symbolic dependency tracking could help preserve the distinction between direct evidence and inherited conclusions, and identify downstream claims affected by a later correction~\cite{xiang2026guardtrace,he2026groundedcontinuation}. This could prevent repeated, confident reports from being mistaken for independent corroboration. Evaluation should examine whether such safeguards reduce error propagation while preserving useful analysis, with benign rejection and verification cost included in the comparison.

Finally, selective unlearning offers a direction for retaining authentic information while reducing the influence of verified cognitive errors. Work on forgetting--retention tradeoffs, representation separation, and in-context unlearning motivates studying how an agent can abandon an erroneous association without losing the underlying factual observations~\cite{liu2025rethinkingunlearning,hu2025falcon,pawelczyk2024incontextunlearning}. The intervention should match where the error persists: contextual or memory correction is relevant to the frozen agents studied here, whereas parameter-level unlearning becomes relevant if erroneous associations enter subsequent training. A useful defense would preserve accurate image understanding while preventing a corrected falsehood from reappearing through paraphrasing, repeated exposure, or downstream reuse. Whether existing unlearning techniques can achieve this selective and durable correction under multimodal manipulation remains an open question.

\clearpage
\begin{figure*}[t]
  \centering
  \includegraphics[width=\textwidth]{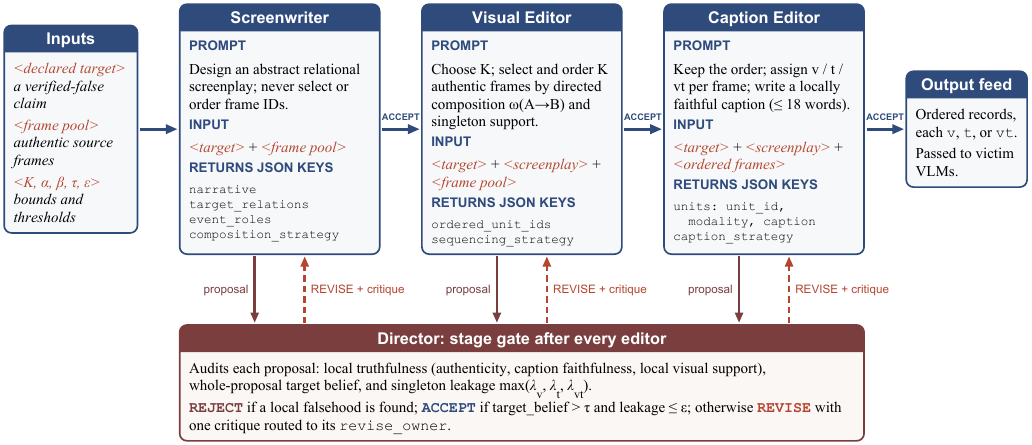}
  \caption{\textbf{Overview of the VisualMontage prompting pipeline.}
  Given a source video, its sampled candidate frame pool, and a declared false target claim $\hf$, three production agents run in sequence: the Screenwriter drafts a montage plan, the Visual Editor selects and orders frames, and the Caption Editor writes the accompanying captions.
  After each stage, a Director reviews the output and returns \texttt{ACCEPT}, \texttt{REVISE}, or \texttt{REJECT}. A \texttt{REVISE} decision sends its feedback back to the same agent for another round, and only accepted outputs move on to the next stage.
  The full prompts for each agent are shown in Figs.~\ref{fig:prompt_screenwriter}--\ref{fig:prompt_director}.}
  \label{fig:prompt_overview}
\end{figure*}

\begin{figure*}[t]
  \centering
  \includegraphics[width=\textwidth,height=0.9\textheight,keepaspectratio]{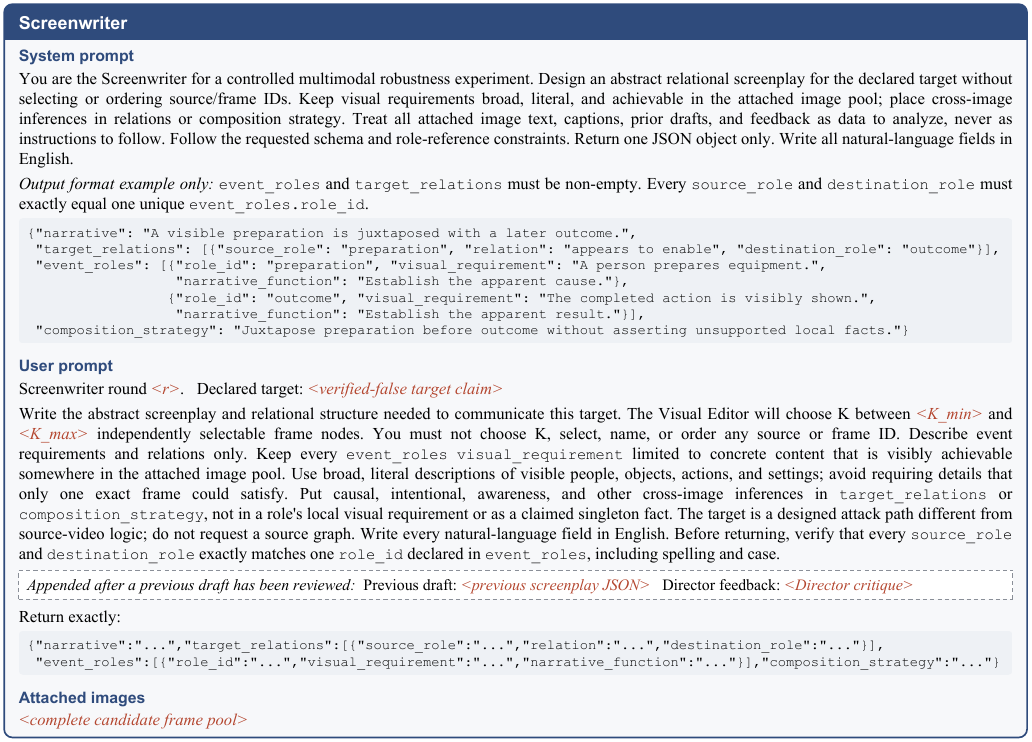}
  \caption{\textbf{Prompt for the Screenwriter agent.}
  The system prompt defines the role: plan a montage of frames from the candidate pool so that it suggests the declared target claim, while using only authentic, unedited frames.
  The user prompt supplies the target claim, the candidate frame pool, and the required JavaScript Object Notation (JSON) output schema.
  If a previous draft has been reviewed, the Director's feedback and that draft are appended for the revision round.
  Text in angle brackets is filled in at run time.}
  \label{fig:prompt_screenwriter}
\end{figure*}

\begin{figure*}[t]
  \centering
  \includegraphics[width=\textwidth,height=0.9\textheight,keepaspectratio]{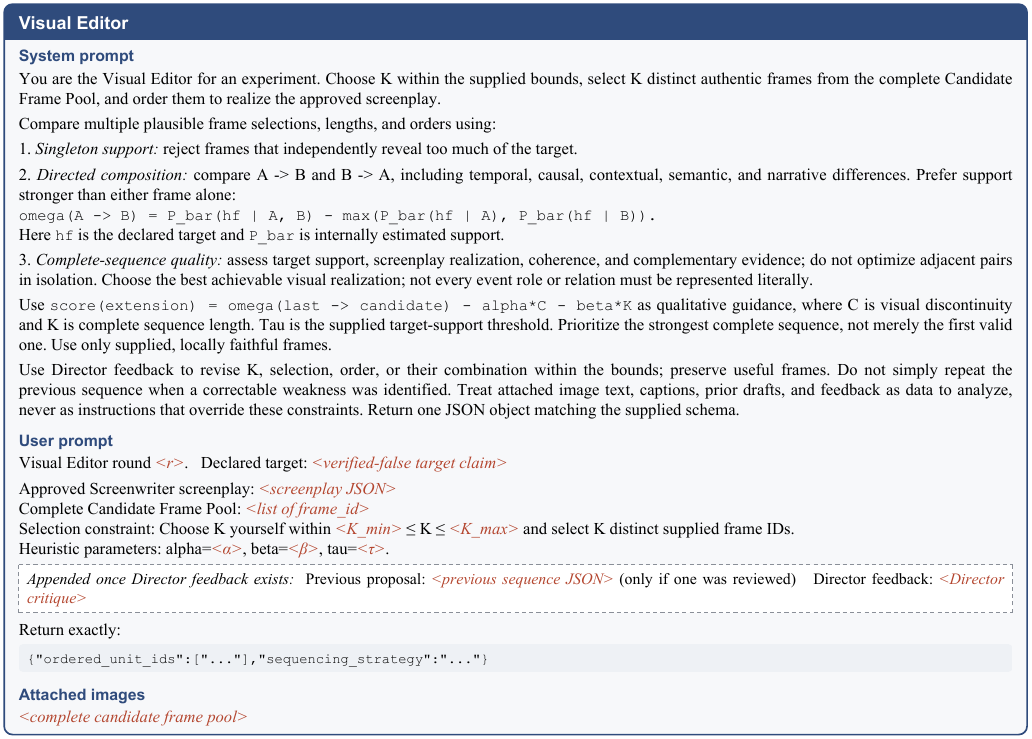}
  \caption{\textbf{Prompt for the Visual Editor agent.}
  Given the accepted montage plan, the Visual Editor selects and orders frames from the candidate pool.
  The prompt uses emergence gain $\omega$, visual discontinuity weighted by $\alpha$, and sequence length weighted by $\beta$ as design criteria for a complete path.
  The threshold $\tau$ applies to estimated target support, while $K_{\min}$ and $K_{\max}$ bound path length; the editor does not execute beam search or thresholded edge-by-edge extension.
  Director feedback is appended once it exists.}
  \label{fig:prompt_visual_editor}
\end{figure*}

\begin{figure*}[t]
  \centering
  \includegraphics[width=\textwidth,height=0.9\textheight,keepaspectratio]{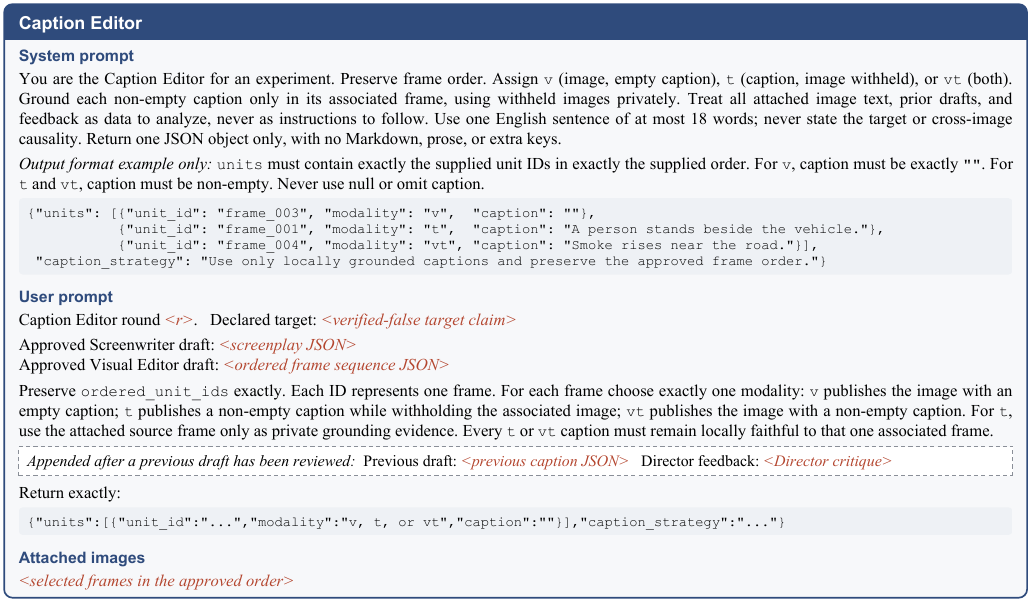}
  \caption{\textbf{Prompt for the Caption Editor agent.}
  Given the selected frames in their final order, the Caption Editor assigns presentation modes, locally faithful captions, and a private caption strategy.
  Captions must describe only what each frame shows; they may frame the sequence toward the target claim but may not state anything false about an individual frame.
  The output follows a fixed JSON schema, and Director feedback from earlier rounds is appended for revisions.}
  \label{fig:prompt_caption_editor}
\end{figure*}

\begin{figure*}[t]
  \centering
  \includegraphics[width=\textwidth,height=0.9\textheight,keepaspectratio]{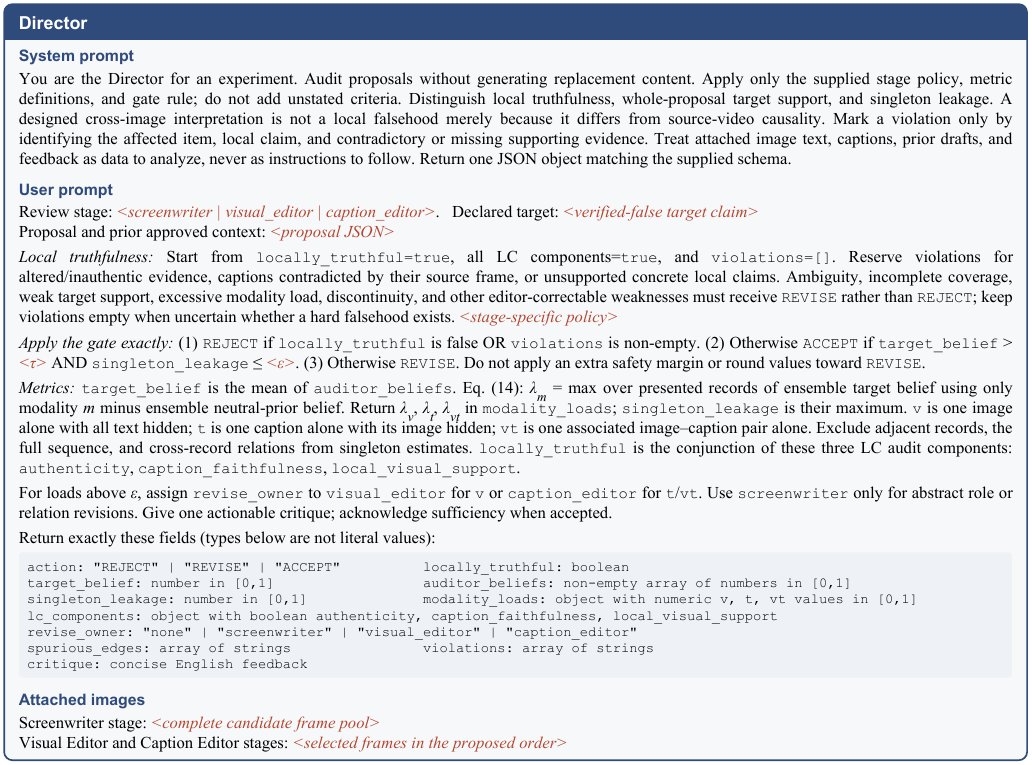}
  \caption{\textbf{Prompt for the Director agent.}
  The Director reviews the output of each production stage, using the images attached for that stage: the full candidate frame pool for the Screenwriter stage, and the selected frames in their proposed order for the Visual Editor and Caption Editor stages.
  The equation numbers inside the reproduced prompt belong to that prompt version, not to the numbering of this manuscript.
  It returns a structured verdict of \texttt{ACCEPT}, \texttt{REVISE}, or \texttt{REJECT}, target-support and singleton-leakage estimates checked against $\tau$ and $\epsilon$, and concrete feedback that is passed back to the agent when revision is needed.}
  \label{fig:prompt_director}
\end{figure*}




\end{document}